\documentclass[10pt,twocolumn,letterpaper]{article}

\usepackage{iccv}              

\definecolor{iccvblue}{rgb}{0.21,0.49,0.74}
\usepackage[pagebackref,breaklinks,colorlinks,allcolors=iccvblue]{hyperref}

\def\paperID{6537} 
\def\confName{ICCV}
\def\confYear{2025}

\usepackage[utf8]{inputenc} 
\usepackage[T1]{fontenc}    
\usepackage{hyperref}       
\usepackage{url}            
\usepackage{booktabs}       
\usepackage{amsfonts}       
\usepackage{nicefrac}       
\usepackage{microtype}      
\usepackage{xcolor}         
\usepackage{graphicx}
\usepackage{relsize}
\usepackage{scalerel}
\usepackage{colortbl}

\usepackage{tikz}
\usepackage{multirow}
\usepackage{multicol}
\definecolor{gold}{HTML}{FBF2D2}
\definecolor{silver}{HTML}{DDDDDD}
\definecolor{bronze}{HTML}{EED2B8}

\definecolor{goldD}{HTML}{D9AE13}
\definecolor{silverD}{HTML}{909090}
\definecolor{bronzeD}{HTML}{9A5F26}

\definecolor{catGreen}{HTML}{238763}
\definecolor{catBlue}{HTML}{1F70AE}

\newcommand{\medal}[3]{\tikz[baseline=(char.base)]{\node[rounded corners=2pt,fill=#1,draw=#2,inner sep=1.5pt] (char) {#3};}}

\newcommand{\bm}[2]{
    \ifcase#1\or
      {\medal{gold}{goldD}{\textbf{#2}}}
    \or 
      {\medal{silver}{silverD}{#2}}
    \or 
      {\medal{bronze}{bronzeD}{#2}}
    \else 
      #2
    \fi\ignorespaces
}

\NewDocumentCommand\salad{}{\scalerel*{\includegraphics{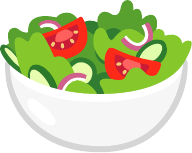}}{X}}

\title{SALAD \salad\, -- Semantics-Aware Logical Anomaly Detection}

\author{Matic Fučka \and Vitjan Zavrtanik \and Danijel Skočaj \and
University of Ljubljana, Faculty of Computer and Information Science, Slovenia\\
{\tt\small \{matic.fucka, vitjan.zavrtanik, danijel.skocaj\}@fri.uni-lj.si}
}

\begin{document}
\maketitle
\begin{abstract}
Recent surface anomaly detection methods excel at identifying structural anomalies, such as dents and scratches, but struggle with logical anomalies, such as irregular or missing object components. The best-performing logical anomaly detection approaches rely on aggregated pretrained features or handcrafted descriptors (most often derived from composition maps), which discard spatial and semantic information, leading to suboptimal performance. We propose SALAD, a semantics-aware discriminative logical anomaly detection method that incorporates a newly proposed composition branch to explicitly model the distribution of object composition maps, consequently learning important semantic relationships. Additionally, we introduce a novel procedure for extracting composition maps that requires no hand-made labels or category-specific information, in contrast to previous methods. By effectively modelling the composition map distribution, SALAD significantly improves upon state-of-the-art methods on the standard benchmark for logical anomaly detection, MVTec LOCO, achieving an impressive image-level AUROC of 96.1\%.
Code: \textcolor{magenta}{\url{https://github.com/MaticFuc/SALAD}}
\end{abstract}    
\section{Introduction}

\begin{figure}
    \centering
    \includegraphics[width=\columnwidth]{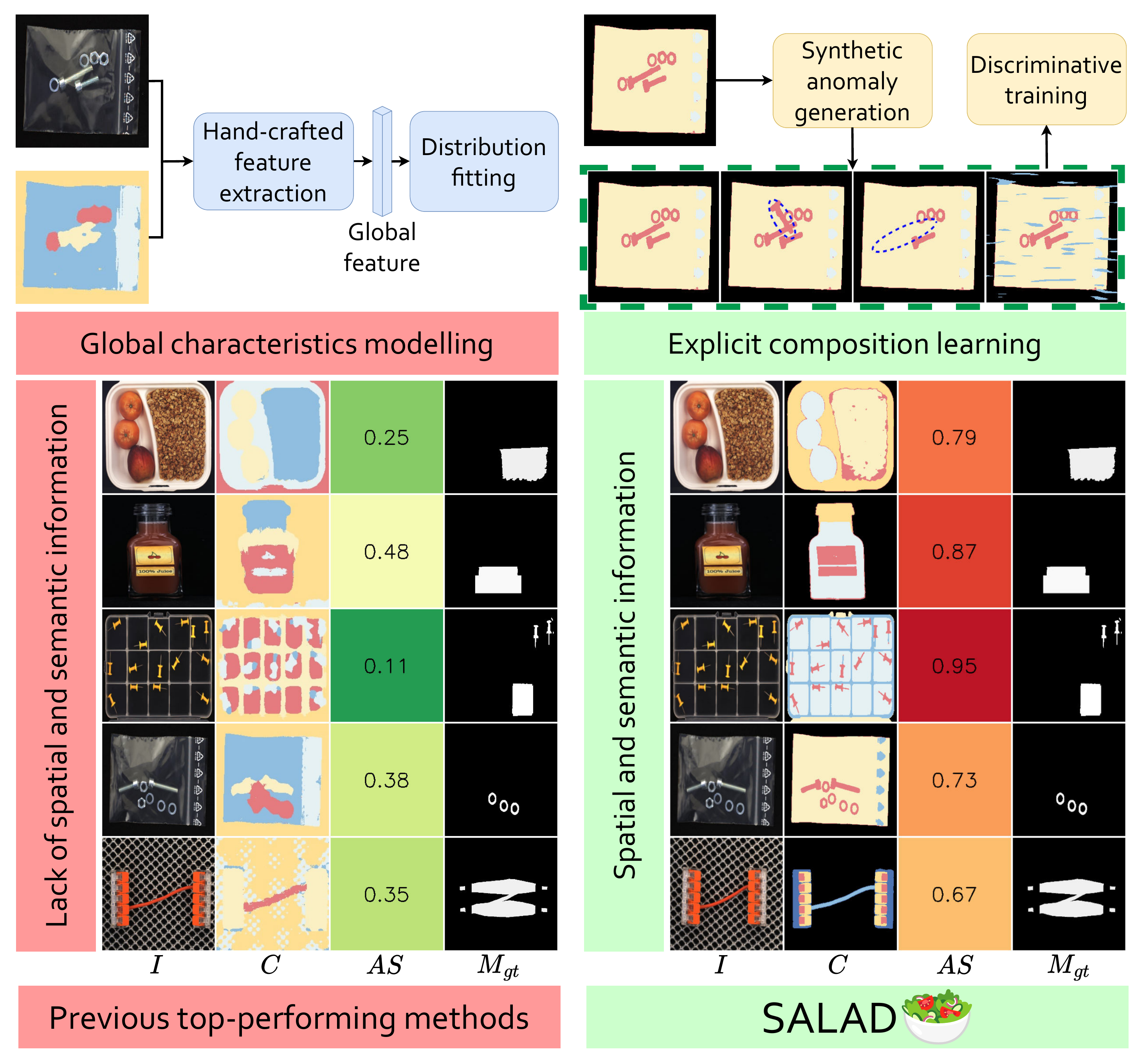}
    \caption{While previous approaches rely on handcrafted features, SALAD explicitly models the composition distribution by training directly on the composition maps $C$. SALAD is trained on simulated anomalous examples that are enclosed by a \textcolor{ForestGreen}{green} dashed rectangle, whereas individual synthetic logical anomalies are highlighted with \textcolor{blue}{blue} ellipses. SALAD improves the estimated anomaly scores ($AS$) on near-in-distribution logical anomalies.
    }
    \label{fig:idea}
\end{figure}

Surface anomaly detection aims to detect and localise abnormal regions in the image while training only on anomaly-free images. It is commonly used in the industrial inspection domain~\cite{mvtec,visa,KSDD2,mvtec-loco, Jakob_PAA} where the limited availability and considerable diversity of abnormal images make training supervised models impractical. The problem of surface anomaly detection can be split into structural and logical anomaly detection. Structural anomalies are irregularities in the local appearance distribution, e.g., dents or scratches. They can be detected by modelling the object's anomaly-free appearance and detecting local texture deviation. Logical anomalies break the semantic constraints of the image, e.g. incorrect number or misplacement of object components. For such anomalies, a model of local appearance does not suffice since object components may fit the anomaly-free appearance distribution locally. Recent top-performing surface anomaly detection methods~\cite{draem,patchcore,simplenet,dsr,transfusion,reverse_dist,cflow-ad,fair} are focused mostly on detecting structural anomalies and fail to model the semantic information required for logical anomaly detection.

Recent logical anomaly detection methods consist of a base structural branch that is typically a time-tested surface anomaly detection method, and a logical anomaly branch focusing on detecting semantic anomalies. The output scores of both branches are fused to produce a final anomaly score. Best performing logical anomaly methods attempt to detect deviations in appearance caused by logical anomalies by either modelling the global image appearance~\cite{puad} or by utilising object composition maps~\cite{comad,psad,csad}. 
Global appearance-based approaches build a global image descriptor by aggregating pretrained features, disregarding the information contained in the object components' position, orientation, and frequency. This results in missed detections in harder, near-in-distribution anomalous examples (Figure~\ref{fig:idea}, left). Additionally, such approaches do not decouple logical and structural anomalies because they use pretrained features that focus on modelling appearance, leading to a considerable emphasis on structural anomalies. Recent composition map-based logical anomaly detection methods extract composition maps from the input RGB images. They then rely on handcrafted features extracted from composition maps, such as the class frequency, to better model the global distribution. Similar to the global appearance approaches, such representations do not sufficiently model the available semantic and spatial information. Additionally, the best-performing composition-based methods require either hand-labelled training examples~\cite{psad} or category-specific procedures~\cite{csad} to extract composition maps, making them infeasible to apply to new datasets.

We hypothesise that training a model (in our case, a composition branch) to model the composition map distribution would also capture the critical spatial and semantic information unobtainable with global representations, which would improve logical anomaly detection performance. Discriminative methods, which rely on synthetic anomalies to learn an anomaly-free distribution, present a possible solution. However, synthetic anomalies are currently defined only for RGB images~\cite{draem, memseg}. Therefore, we adapted them for composition maps. Additionally, as composition maps contain a compressed representation of object class, shape, and position while discarding appearance information, it is simpler to simulate more expressive anomalies appropriate for detecting logical anomalies. We propose a composition branch defined as a discriminative anomaly detection model operating with composition maps (Figure~\ref{fig:idea}, right).

The contributions of this work are twofold. \textit{(i)} As our main contribution, we propose SALAD, a \textbf{S}emantics-\textbf{A}ware discriminative \textbf{L}ogical \textbf{A}nomaly \textbf{D}etection method that extends the recent appearance and global branch framework with a new compositional branch that explicitly learns the anomaly-free object composition distribution. \textit{(ii)} As an additional contribution, we propose a novel object composition map generation process. The proposed approach produces maps of high quality (Figure~\ref{fig:idea}, right, columns $C$) without requiring hand-labelled training data or category-specific procedures in contrast to previous approaches. We showcase its robustness by applying it to several objects and datasets. 

To emphasise the value of the proposed contributions, extensive experiments are performed on the standard logical anomaly detection benchmark, MVTec LOCO~\cite{mvtec-loco}. SALAD achieves a new state-of-the-art result on MVTec LOCO (AUROC of 96.1\%), outperforming competing methods by a significant margin of $3.0$ percentage points. Additionally, SALAD is evaluated on standard MVTec AD~\cite{mvtec} and VisA~\cite{visa} datasets, achieving excellent results (AUROC of 98.9\% and 97.9\%).

\section{Related work}

\noindent\textbf{Surface anomaly detection} has been extensively researched, with methods categorised into three main paradigms: reconstructive, embedding-based, and discriminative.

\textit{Reconstructive methods} train either an autoencoder-like~\cite{fair,ZavrtanikInpainting} network, a generative adversarial network~\cite{ocrgan}, a diffusion model~\cite{AnnoDDPM} or a transformer~\cite{intra,intracorr} on anomaly-free images and assume the resulting model will not generalise well to anomalous regions, making them distinguishable by reconstruction error. 
\textit{Embedding-based methods} leverage pretrained models to extract features and fit a normality model on top of them. The normality methods are often a coreset~\cite{patchcore}, a student-teacher network~\cite{reverse_dist,efficientad,ast} or a normalising flow network~\cite{cflow-ad,fastflow}. 
\textit{Discriminative methods} focus on learning the boundary between normal and abnormal samples. Methods in this paradigm are learned to segment synthetic anomalies~\cite{draem,simplenet,dsr,transfusion,memseg,supersimplenet} and learn a normality model to generalise to real-world scenarios.
These methods fail on logical anomalies as they focus on local characteristics and do not consider global semantics.

\noindent\textbf{Logical anomaly detection} is a new surface anomaly detection research direction. The methods can be divided into three paradigms: local-global reconstruction, global distribution approaches and composition-based.

\textit{Local-global reconstruction methods} use a two-branch neural network~\cite{mvtec-loco,efficientad,thfr,reverse_dist_logical,grad,slsg}. These approaches contain a local and global appearance branch and assume that structural anomalies occur as local deviations and logical anomalies impact a large part of the image, which does not always hold. EfficientAD~\cite{efficientad} uses a convolutional network with a small receptive field as the local branch and an autoencoder as the global branch. These methods have difficulties with categories without a constant object layout.

\textit{Global distribution approaches} extract global appearance descriptors from images and use a descriptor distribution model~\cite{sinbad, samlad} to detect anomalies. LogicAD~\cite{logicad} uses a large Vision Language Model to extract a global description. This is converted into a formal representation and evaluated by an automatic theorem prover. PUAD~\cite{puad} estimates the global distribution by fitting a Gaussian with mean feature vectors extracted from feature maps produced by EfficientAD's student. During inference, the anomaly is detected using Mahalanobis distance. Simply using the mean of extracted pretrained features as a global appearance representation disregards considerable spatial information, leading to poor performance on spatially dependent logical anomalies. Introducing spatial information might improve the logical anomaly detection performance.

\begin{figure}[t]
    \centering
    \includegraphics[width=1\columnwidth]{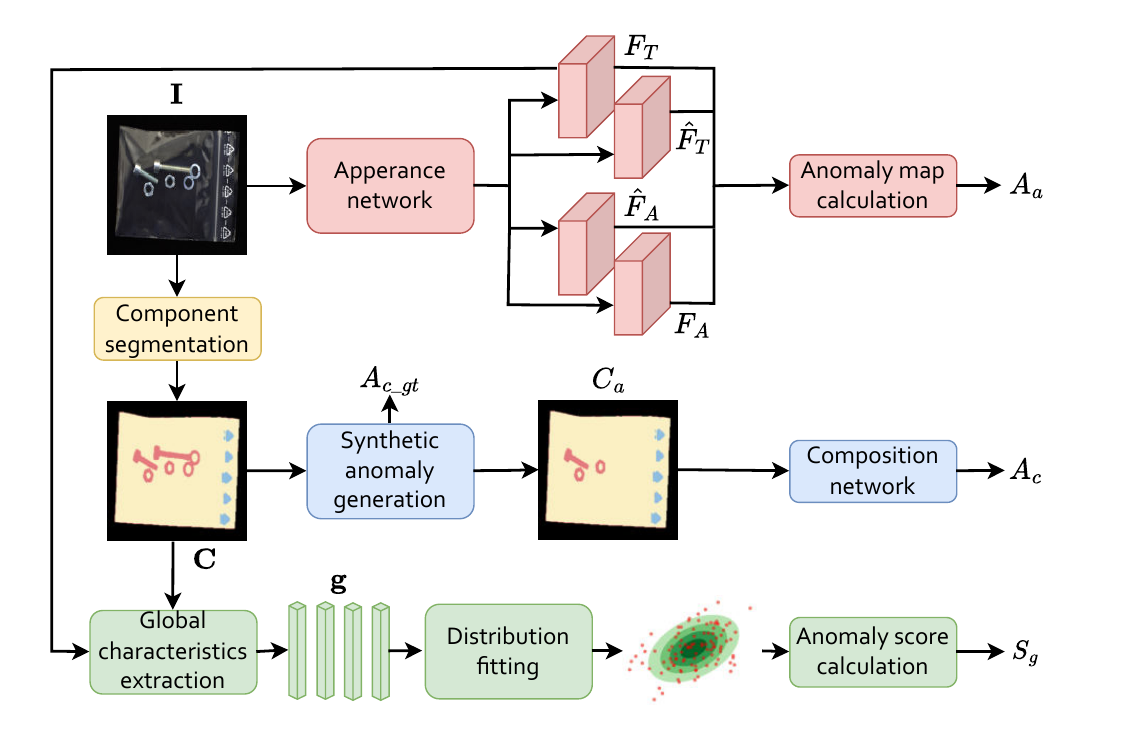}
    \caption{SALAD is constructed from a \textcolor{red}{local appearance branch}, a \textcolor{blue}{composition branch}, and a \textcolor{ForestGreen}{global branch}. Each branch focuses on a different level of image semantics. Synthetic anomalies are generated to train the composition branch. Composition maps are segmented using a \textcolor{Goldenrod}{component segmentation network}.}
    \label{fig:overview}
\end{figure}

\textit{Composition-based approaches} model the semantic information by using composition maps, i.e. object component segmentation maps of the image. This paradigm was introduced with ComAD~\cite{comad}, which extracts features from a pretrained network and clusters them to create a rough semantic segmentation. The obtained segmentation maps are used only to extract handcrafted features and store them in a memory bank. PSAD~\cite{psad} uses hand-labelled segmentation maps to finetune a pretrained feature extractor with an attached segmentation head. After the fine-tuning, PSAD creates a memory bank with global statistics and extracted feature vectors. PSAD's manual annotation requirement is impractical in real-world scenarios. CSAD~\cite{csad} obtains patch histograms of composition maps and stores them in a memory bank. CSAD extracts object composition maps automatically, but the parameters for each object category must be tuned, which is a drawback in practical use.

Unlike recent logical anomaly detection methods, SALAD does not focus on feature averaging or handcrafted features but explicitly models the composition map distribution, thus learning important composition information.
We also propose a highly accurate composition map generation procedure that does not require hand-labelled data or category-specific information.

\section{SALAD}

Recent logical anomaly detection methods are composed of a base local appearance model and a global model. The global model typically constructs the global distribution by either using aggregated pretrained features or handcrafted descriptors. The best-performing methods use composition maps to create better global representations. Due to such a construction, a significant amount of semantic information is not captured. In the proposed method, SALAD (Figure~\ref{fig:overview}), we follow the initial framework of a base structural anomaly detection model with a global appearance branch but propose a novel composition branch that explicitly learns the distribution of composition maps, consequently learning critical spatial and semantic information. Additionally, SALAD contains an improved global appearance model.

In the composition branch, the anomaly-free object composition distribution is learned in a discriminative fashion through a novel anomaly simulation process. It uses the information-dense composition maps to simulate near-in-distribution anomalies that are difficult to simulate using traditional simulation processes~\cite{draem}. Such simulated anomalies facilitate the learning of a tight decision boundary around the anomaly-free semantic structure of the images, leading to an improvement in logical anomaly detection performance.

At inference time, structural anomalies are detected by the local appearance branch, while the global appearance and the composition branch focus on logical anomalies. Individual branch outputs are then combined using a score fusion model. In the rest of this section, we describe SALAD in detail.

\begin{figure}[t]
    \centering
    \includegraphics[width=1\columnwidth]{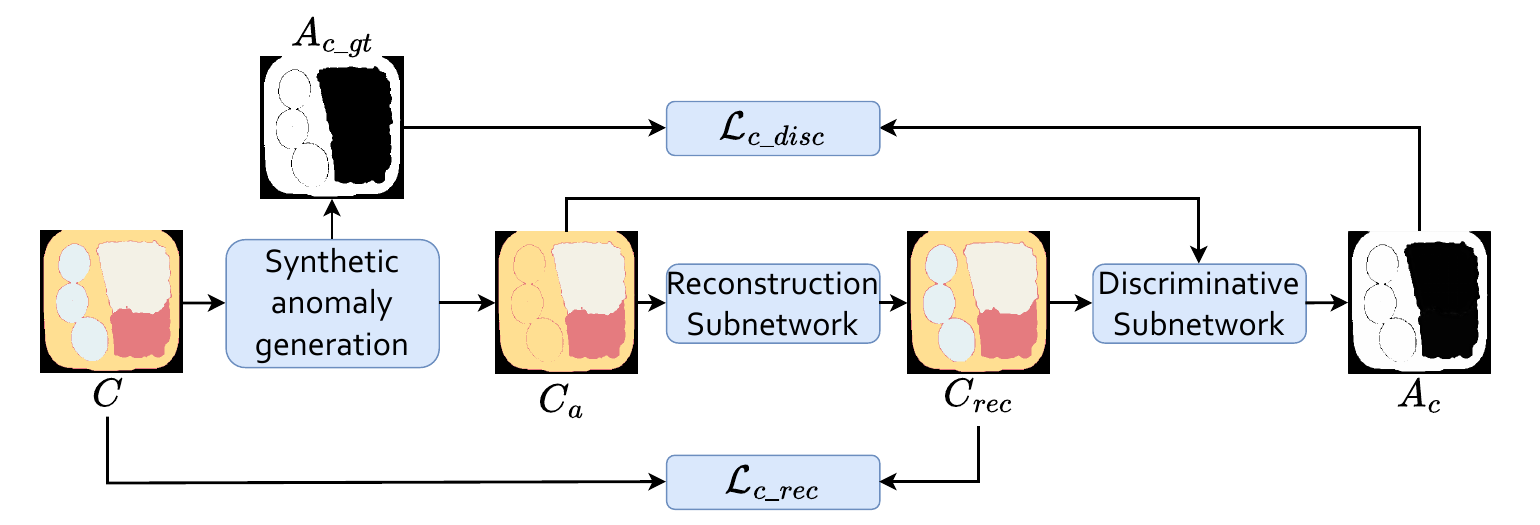}
    \caption{Composition branch architecture.
    }
    \label{fig:comp_branch}
\end{figure}

\begin{figure*}[t]
    \centering
    \includegraphics[width=\textwidth]{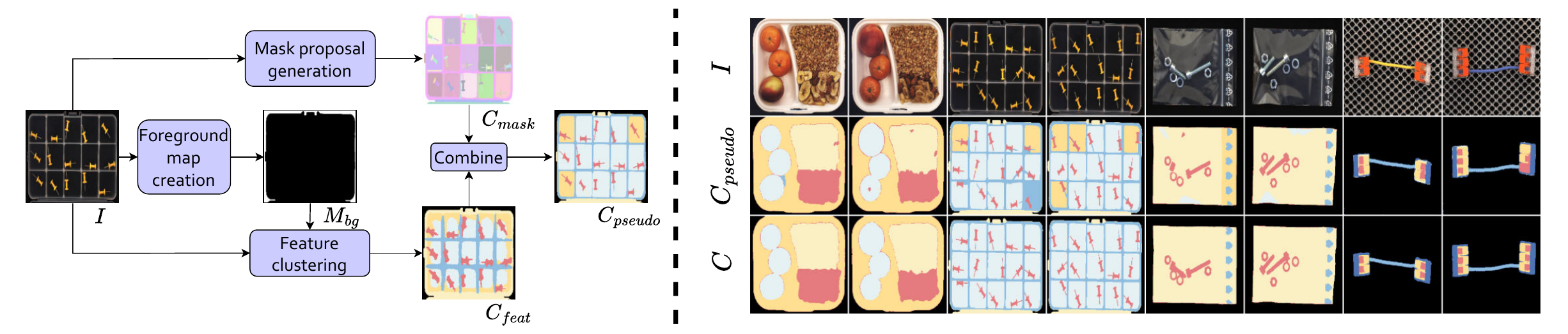}
    \caption{Left: Pseudo label $C_{pseudo}$ generation procedure. The main idea is to combine precise mask generation from SAM-HQ~\cite{sam_hq} and the discriminative power from DINO~\cite{dino}. 
    Right: Comparisons of the final object composition maps $C$ (generated by the \textit{composition segmentation model}) and the pseudo labels $C_{pseudo}$. Input images $I$ are also depicted for comparison. Final object composition maps $C$ contain less misclassified components than the pseudo labels $C_{pseudo}$.
    }
    \label{fig:noisy_component_map}
\end{figure*}

\subsection{Local appearance model} 
Following recent advancements in logical anomaly detection, a powerful surface anomaly detection model is used as the base structural anomaly detection branch. In SALAD, the initial structural branch follows the EfficientAD architecture~\cite{efficientad}, a top-performing surface anomaly detection framework. EfficientAD takes an RGB image $I$ as input and works as an embedding reconstruction model. The input image $I$ is mapped to a feature representation $F_T$ by a teacher encoder trained on natural images~\cite{imagenet}. An autoencoder network takes $I$ as input and is tasked with reconstructing the teacher output $F_T$, outputting $F_A$. The student encoder is tasked with reconstructing both $F_T$ and $F_A$, outputting $\hat{F}_T$ and $\hat{F}_A$. The anomaly map $A_a$ output by EfficientAD is based on the difference between $F_T$ and $\hat{F}_T$ and between $F_A$ and $\hat{F}_A$.

\subsection{Object composition model}

The composition branch is formulated as a discriminative anomaly detection method that operates on object composition maps $C$. Training such a model requires both an automated way of obtaining object composition maps (Section \ref{sec:composition_map}) as well as a well-defined anomaly simulation process specifically designed for composition maps (Section \ref{sec:synthetic}). 

The \textit{composition branch} follows a general discriminative anomaly detection architecture~\cite{draem} composed of a composition reconstruction network and a composition discriminative network (Figure \ref{fig:comp_branch}). The composition reconstruction network first restores the (synthetically) anomalous parts of the composition map to their anomaly-free appearance. Then, the input composition map and its anomaly-free reconstruction are passed to the composition discriminative network to output an anomaly mask. The composition branch operates solely in the space of object composition maps $C$. During training, the composition reconstruction network takes as input a composition map $C_{a}$, which has been augmented to include simulated anomalies. The network is then trained to restore the original anomaly-free composition $C$ by outputting an anomaly-free composition reconstruction $C_{rec}$. During inference, $C$ is used as the input instead of $C_a$.

Since the composition map $C$ values belong to individual classes, segmentation losses are used to train the composition reconstruction network. Namely, the focal loss $\mathcal{L}_{foc}$~\cite{focal} and dice loss $\mathcal{L}_{dice}$~\cite{dice} are used:
\begin{equation}
    \mathcal{L}_{c\_rec} = \mathcal{L}_{foc}(C, C_{rec}) + \mathcal{L}_{dice}(C, C_{rec}) \enspace ,
\end{equation}
where $C$ is the original object composition map, $C_{rec}$ is the reconstructed object composition map. The composition reconstruction network generalises to real anomalous examples, successfully reconstructing them to be anomaly-free.

After obtaining the anomaly-free composition reconstruction $C_{rec}$, $C_{rec}$ and the augmented composition map $C_a$ (during inference, $C_{rec}$ and $C$ are used) are concatenated and used as input for the composition discriminative network. The discriminative network is trained to predict the difference between $C_{rec}$ and $C_a$ to output an anomaly segmentation map $A_{c}$.

Following recent literature~\cite{memseg}, the loss for the composition discriminative network is defined as:
\begin{equation}
    \mathcal{L}_{c\_disc} = \alpha\mathcal{L}_{foc}(A_{c\_gt}, A_{c}) + \mathcal{L}_1(A_{c\_gt}, A_{c}) \enspace ,
\end{equation}
where $A_{c\_gt}$ is the ground truth anomaly map corresponding to synthetic anomalies, $A_{c}$ is the predicted anomaly map, and $\alpha$ is the weighting parameter (set to 5 in all our experiments).

\subsection{Object composition maps} \label{sec:composition_map}

A semantically meaningful representation of the composition must first be extracted to model the object composition distribution accurately. Segmentation of object parts concisely represents part frequency, shapes, sizes, and positions without additional appearance information that is redundant for object composition and increases the complexity of the representation. Accurate component-level segmentation maps (dubbed composition maps) are used in SALAD.

A two-step process is used for composition map extraction. First, \textit{pseudo-labels} $C_{pseudo}$ for the training set are created by clustering DINO~\cite{dino} features to obtain rough segmentation maps $C_{feat}$, which are then used to classify highly accurate mask proposals $C_{mask}$ generated by SAM-HQ~\cite{sam_hq}. Finally, the pseudo-labels are used to train a \textit{component segmentation model} to predict the final object composition maps $C$. Pseudo-label creation is illustrated in Figure~\ref{fig:noisy_component_map}.

For the initial pseudo-labels, background maps for each training image are generated by querying SAM-HQ on image corners and combining the resulting masks to produce a background mask. This mask is inverted to create the foreground mask $M_{fg}$. DINO~\cite{dino} feature maps are extracted and resized to $256\times256$. Features outside $M_{fg}$, that is, the background features, are set to $0$ to reduce noise; the rest are then subsampled and clustered into $K$ clusters (in our case $K$=6) to produce a rough object composition map $C_{feat}$.

SAM-HQ is queried on a grid over the input image $I$ to obtain mask proposals $C_{mask}$. Each mask proposal is classified as the class of the corresponding majority cluster in $C_{feat}$, aligning high-quality masks with component labels to create high-quality pseudo-labels $C_{pseudo}$. Due to the computational intensity of SAM-HQ and DINO, a \textit{component segmentation model} is trained with $I$ and corresponding $C_{pseudo}$ pairs. Specifically, we use a simple UNet trained with a cross-entropy loss. Even though some $C_{pseudo}$ contain mistakes, the components are correctly classified on average across the dataset. Due to that, the component segmentation model generalises and outputs composition maps $C$ without incorrectly labelled components. The trained component segmentation model infers the desired composition map $C$ directly from $I$, enabling efficient composition map extraction. Figure~\ref{fig:noisy_component_map} shows examples of $C_{pseudo}$ and object composition maps $C$ produced by the component segmentation model.

\subsection{Synthetic anomaly generation} \label{sec:synthetic}

An appropriate synthetic anomaly generation procedure is required to facilitate the training of the discriminative composition branch. Due to differences between structural and logical anomalies, different anomaly generation strategies are required. To simulate structural anomalies, we extend the synthetic anomaly generation proposed by DR{\AE}M~\cite{draem} by pasting a random class on top of the composition map according to an anomaly map generated using Perlin noise~\cite{perlin1985image}. To generate near-in-distribution logical anomalies, a random component is either inpainted (from another image) or erased. When a component is erased, the corresponding region is inpainted with a component class randomly selected from the neighbouring components. In this case, the anomaly map marks both the erased component and the neighbouring component class used for inpainting. For cases where a component is inpainted, identifying the exact anomaly location is ill-posed (e.g., an extra screw added to a screw bag). Therefore, the anomaly map in these cases includes all regions containing the inpainted component class. Several examples of synthetic anomalies can be seen in Figure~\ref{fig:idea}.

\subsection{Global appearance model} 
Using a strong global appearance model can also improve the detection of structural anomalies. The global appearance branch utilises features extracted from the input image $I$ and its corresponding object composition map $C$. In $C$, pixels belonging to individual image components are marked with their corresponding class labels $c$. 

For each class label $c$ in $C$, the mean feature vector $g_c$ is computed from the feature vectors in $F_T$ corresponding to pixels belonging to $c$ in $C$. The set of $g_c$ values for all classes $C$ represents the global appearance descriptor $g$. 

The procedure is repeated for each sample $i$ in the training set to obtain global appearance descriptors $g^{(i)}$ upon which the global distribution is estimated by fitting a Gaussian distribution~\cite{mahalanobis}. The mean $\mu_c$ and covariance $\Sigma_c$ for each class in $C$ are calculated from all samples $g_{c}^{(i)}$ in the training set. During inference, the anomaly score $S_g$ is calculated using the average Mahalanobis distance~\cite{mahalanobis} for each class, that is:
\begin{equation}
    S_g = \frac{1}{K}\sum_{c=1}^K\sqrt{(g_c-\mu_c)^\top \Sigma_c^{-1}(g_c-\mu_c)},
\end{equation}
where $K$ is the total number of classes in $C$. 

\subsection{Anomaly score calculation}

Each model branch outputs an anomaly score at inference: $AS_{a}$, $AS_{c}$ and $AS_{g}$ for the appearance, compositional and global branches, respectively. Individual scores are calculated as follows:
\begin{equation}
\label{eq:final_mask_calc}
AS_{a} = max(A_{a}), \enspace AS_{c} = max(A_{c}), \enspace AS_{g} = S_{g} \enspace ,   
\end{equation}
where $A_a$ is the output of the appearance branch, $A_c$ is the output of the composition branch, and $S_g$ is the output of the global branch. 
The outputs are normalised using the means $\mu_a$, $\mu_c$ and $\mu_g$ and standard deviations $\sigma_a$, $\sigma_c$ and $\sigma_g$ of the anomaly scores on the validation set. The final anomaly score is then defined as:
\begin{equation}
    AS = \frac{AS_{a} - \mu_a}{\sigma_a} + \frac{AS_{c} - \mu_c}{\sigma_c} + \frac{AS_{g} - \mu_g}{\sigma_g} \enspace .
\end{equation}

\begin{table*}[t]
\centering
\resizebox{\textwidth}{!}{
    \begin{tabular}{lcccccccc}
\toprule
\textbf{Method} & \textbf{Venue} & \textbf{Supervised Masks} & Breakfast box & Juice bottle & Pushpins & Screw bag & Splicing conn. & \textit{Average} \\ \midrule 
DR{\AE}M~\cite{draem} & ICCV'21 & & 80.2 & 94.3 & 68.6 & 70.6 & 85.4 & 79.8  \\
TransFusion~\cite{transfusion} & ECCV'24 & & 82.4 & \bm1{99.7} & 63.8 & 71.5 & 83.7 & 80.2  \\
DSR~\cite{dsr} & ECCV'22 & & 85.8 & 99.2 & 76.5 & 64.9 & 85.5 & 82.6  \\
THFR~\cite{thfr} & ICCV'23 & & 77.3 & 80.1 & 80.8 & 79.7 & 81.0 & 83.3  \\
LogicAD~\cite{logicad} & AAAI'25 & & \bm1{92.1} & 81.6 & \bm2{98.1} & 83.8 & 73.4 & 86.0  \\
Sinbad~\cite{sinbad} & ArXiv'23 & & \bm2{91.8} & 94.4 & 83.9 & \bm2{86.8} & 84.5 & 88.3  \\
ComAD + Patchcore~\cite{comad} & AEI'23 & & 86.4 & 96.6 & 93.4 & 80.2 & 94.1 & 90.1  \\
SLSG~\cite{slsg} & PR'24 & & 88.9 & 99.1 & 95.5 & 79.4 & 88.5 & 90.3  \\
SAM-LAD~\cite{samlad} & KBS'25 & & \bm3{91.0} & 97.6 & 88.2 & \bm3{86.6} & 90.0 & \bm3{90.7}  \\
EfficientAD~\cite{efficientad} & WACV'24 & & 88.5 & 99.0 & 93.6 & 73.6 & \bm2{97.1} & \bm3{90.7}  \\
PUAD~\cite{puad} & ICIP'24 & & 87.1 & \bm1{99.7} & \bm3{98.0} & 81.1 & \bm3{96.8} & \bm2{93.1}  \\

\rowcolor{gray!10}\textbf{\textit{SALAD}} &  & & 89.3 & \bm1{99.7} & \bm1{99.4} & \bm1{95.0} & \bm1{97.3} & \bm1{96.1}  \\ \midrule

PSAD~\cite{psad} & AAAI'24 & \checkmark & \bm3{92.5} & \bm2{98.7} & \bm3{94.9} & \bm1{97.5} & \bm3{90.6} & \bm3{94.9}  \\
CSAD~\cite{csad} & BMVC'24 & \checkmark & \bm2{92.8} & \bm3{95.3} & \bm2{98.7} & \bm3{96.5} & \bm2{93.5} & \bm2{95.3}  \\

\rowcolor{gray!10} \textbf{\textit{SALAD}}$^\dagger$ &  & \checkmark & \bm1{94.2} & \bm1{99.3} & \bm1{99.1} & \bm2{96.6} & \bm1{97.0} & \bm1{97.2}  \\
 \bottomrule 

    \end{tabular}
}
\caption{Anomaly detection (AUROC) on MVTec LOCO~\cite{mvtec-loco}. \textcolor{goldD}{First}, \textcolor{silverD}{second} and \textcolor{bronzeD}{third} place are marked. \textit{SALAD}$^\dagger$ is trained using composition maps from PSAD~\cite{psad}.}
\label{tb:results} 
\end{table*}

\begin{table*}[t]
\centering
\resizebox{\textwidth}{!}{
    \begin{tabular}{lcccccccccccccc}
\toprule
\multirow{2}{*}{\textbf{Category}} & \multirow{2}{*}{\textbf{Venue}} & \multirow{2}{*}{\textbf{Supervised Masks}} & \multicolumn{2}{c}{Breakfast box} & \multicolumn{2}{c}{Juice bottle} & \multicolumn{2}{c}{Pushpins} & \multicolumn{2}{c}{Screw bag} & \multicolumn{2}{c}{Splicing conn.} & \multicolumn{2}{c}{\textit{Average}} \\
 & & & Log. & Str. & Log. & Str. & Log. & Str. & Log. & Str. & Log. & Str. & Log. & Str. \\  \midrule 
DR{\AE}M~\cite{draem} & ICCV'21 & & 75.1 & 85.4 & 97.8 & 90.8 & 55.7 & 81.5 & 56.2 & 85.0 & 75.2 & 95.5 & 72.0 & 87.6  \\
TransFusion~\cite{transfusion} & ECCV'24 & & 78.8 & 86.0 & \bm1{99.8} & \bm1{99.6} & 56.4 & 71.3 & 54.8 & 88.2 & 69.2 & \bm2{98.2} & 71.8 & 88.7  \\
DSR~\cite{dsr} & ECCV'22 & & 83.6 & \bm3{88.0} & \bm3{99.5} & 98.9 & 69.4 & 83.6 & 54.4 & 75.4 & 75.9 & 94.9 & 75.0 & 90.2  \\
Sinbad~\cite{sinbad} & ArXiv'23 & & \bm1{97.7} & 85.9 & 97.1 & 91.7 & 88.9 & 78.9 & 81.1 & \bm2{92.4} & 91.5 & 78.3 & \bm3{91.2} & 85.5  \\
ComAD + Patchcore~\cite{comad} & AEI'23 & & 81.6 & \bm1{91.1} & 98.2 & 95.0 & 91.1 & \bm2{95.7} & \bm3{88.5} & 71.9 & \bm3{94.9} & 93.3 & 89.4 & 90.9  \\
SLSG~\cite{slsg} & PR'24 & & \bm3{93.7} & 84.5 & 99.2 & 98.8 & \bm2{97.4} & 93.4 & 69.4 & \bm3{91.6} & 88.4 & 88.5 & 89.6 & \bm3{91.4}  \\
SAM-LAD~\cite{samlad} & KBS'25 & & \bm2{96.7} & 85.2 & 98.7 & 96.5 & \bm3{97.2} & 79.2 & \bm1{95.2} & 77.9 & 91.4 & 88.6 & \bm2{95.8} & 85.5  \\
EfficientAD~\cite{efficientad} & WACV'24 & & 87.4 & \bm2{89.5} & 98.8 & \bm3{99.1} & 93.5 & \bm3{93.6} & 58.1 & 89.1 & \bm1{96.0} & \bm2{98.2} & 86.8 & \bm2{94.7}  \\
\rowcolor{gray!10} \textbf{\textit{SALAD}} &  & & 92.9 & 85.7 & \bm2{99.7} & \bm1{99.6} & \bm1{100.0} & \bm1{98.8} & \bm2{93.9} & \bm1{96.0} & \bm1{96.0} & \bm1{98.6} & \bm1{96.5} & \bm1{95.7}  \\ \midrule
PSAD~\cite{psad} & AAAI'24 & \checkmark & \bm1{100.0} & \bm3{84.9} & \bm2{99.1} & \bm2{98.2} & \bm1{100.0} & \bm3{89.8} & \bm2{99.3} & \bm1{95.7} & \bm3{91.9} & \bm3{89.3} & \bm2{98.1} & \bm3{91.6}  \\
CSAD~\cite{csad} & BMVC'24 & \checkmark & \bm3{94.4} & \bm1{91.1} & \bm3{94.9} & \bm3{95.6} & \bm3{99.5} & \bm2{97.8} & \bm1{99.9} & \bm3{93.2} & \bm2{94.8} & \bm2{92.2} & \bm3{96.7} & \bm2{94.0}  \\
\rowcolor{gray!10} \textbf{\textit{SALAD}}$^\dagger$ &  & \checkmark & \bm2{99.6} & \bm2{88.8} & \bm1{99.6} & \bm1{98.9} & \bm2{99.9} & \bm1{98.3} & \bm3{98.6} & \bm2{94.7} & \bm1{95.8} & \bm1{98.6} & \bm1{98.7} & \bm1{95.8}  \\
 \bottomrule 
    \end{tabular}
}
\caption{Anomaly detection (AUROC) split by type (Logical/Structural) on MVTec LOCO~\cite{mvtec-loco}.}
\label{tb:results_split}
\end{table*}

\section{Experiments}

\subsection{Datasets}

Experiments are performed on the standard anomaly detection dataset for logical anomalies: the MVTec LOCO~\cite{mvtec-loco} and two standard anomaly detection datasets for structural anomalies: MVTec AD~\cite{mvtec} and VisA~\cite{visa}. The MVTec LOCO dataset comprises 3,644 images distributed across five object categories, the MVTec AD dataset comprises 5,354 images distributed across ten object categories and five texture categories, and the VisA dataset comprises 10,821 images distributed across twelve categories. All three datasets provide pixel-level annotations for the test images, enabling accurate evaluation and analysis.

\subsection{Implementation Details}

\begin{table}[t]
\centering
\resizebox{\columnwidth}{!}{
    \begin{tabular}{lcccc}
\toprule
\textbf{Category} & \textbf{{Logical}} & MVTec AD & VisA & \textit{Average} \\ \midrule 
SimpleNet~\cite{simplenet} &  & \bm1{99.6} & 87.9 & 93.8  \\
DR{\AE}M~\cite{draem} &  & 98.0 & 88.7 & 93.4  \\
TransFusion~\cite{transfusion} &  & \bm2{99.2} & \bm1{98.5} & \bm1{98.9}  \\
DSR~\cite{dsr} &  & 98.2 & 91.6 & 94.9  \\
RD4AD~\cite{reverse_dist} &  & 98.5 & 96.0 & 97.3  \\
Patchcore~\cite{patchcore} &  & \bm3{99.1} & 94.3 & 96.7  \\
EfficientAD~\cite{efficientad} & \checkmark & \bm3{99.1} & \bm2{98.1} & \bm2{98.6}  \\
PUAD~\cite{puad} & \checkmark & 98.5 & 96.9 & 97.7  \\
CSAD~\cite{csad} & \checkmark & 96.2 & 89.5 & 92.6  \\
PSAD~\cite{psad} & \checkmark & 98.0 & 90.3 & 94.2  \\
\rowcolor{gray!10} \textbf{\textit{SALAD}} & \checkmark & 98.8 & \bm3{97.9} & \bm3{98.3}  \\
 \bottomrule 
    \end{tabular}
}
\caption{Anomaly detection (AUROC) on MVTec AD~\cite{mvtec} and VisA~\cite{visa}.}
\vspace{-2cm}
\label{tb:results_mvtec} 
\end{table}

For the composition map generation, SAM-HQ-h~\cite{sam_hq} and a DINO~\cite{dino} pretrained ViT-b$\backslash$8~\cite{vit} are used. A UNet is used as the component segmentation model. The UNet was trained for 15 epochs with AdamW~\cite{adamw} using cross-entropy loss with a learning rate of $5 \cdot 10^{-4}$ and a batch size of $8$. 

SALAD follows the training regime from EfficientAD - 70000 iterations with the Adam~\cite{adam} optimizer. The learning rate was set to $10^{-4}$ for the appearance branch and $10^{-5}$ for the composition branch. Both learning rates were multiplied by $0.1$ after 90\% (66500) of the iterations. Synthetic anomalies were added to the object composition maps with a 50\% probability. All of the images were resized to $256\times256$ pixels. Following the standard protocol, a separate model was trained with a predefined train-test split for each category, and the same hyperparameters were set across all datasets and all categories. Anomaly Scores are normalized using the anomaly scores from the validation set. As MVTec AD and VisA do not have a validation set, we create it by taking a part of the training set (more specifically, 10\%).

\subsection{Experimental results}

\begin{figure*}[t]
    \centering
    \includegraphics[width=1\textwidth]{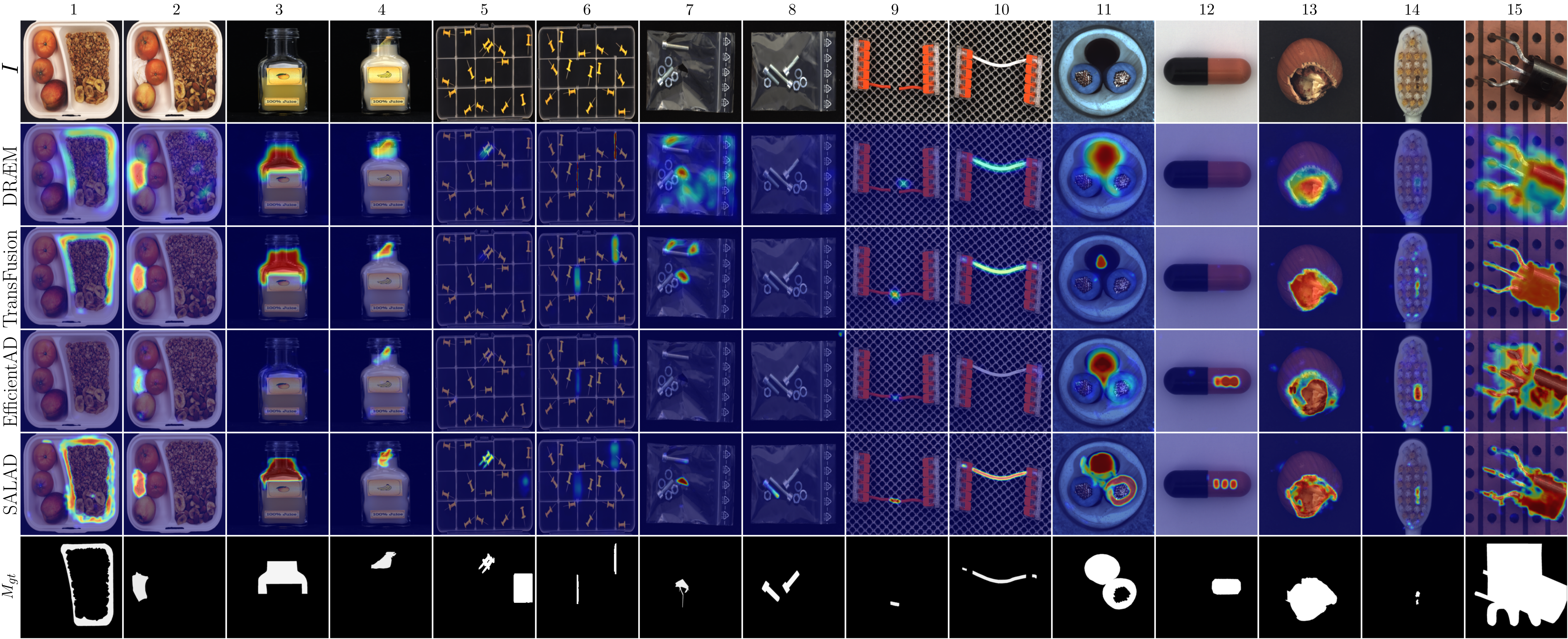}
    \caption{Qualitative comparison of the anomaly segmentation masks produced by SALAD and three other state-of-the-art methods. In the first row, the image is shown. In the next four rows, the anomaly segmentations produced by DR{\AE}M~\cite{draem}, TransFusion~\cite{transfusion}, EfficientAD~\cite{efficientad} and SALAD are depicted, and in the last row, the ground truth mask is depicted. For SALAD, we visualised the sum of $A_a$ and $A_c$ (the outputs of the appearance and the composition branch).}
    \label{fig:res}
\end{figure*}

Following recent literature~\cite{patchcore,efficientad}, anomaly detection performance is evaluated using the Area Under the Receiver Operator Curve (AUROC). Most concurrent works~\cite{puad, psad, csad, sinbad} do not report localisation results due to the ambiguity of the ground-truth masks regarding logical anomalies. Due to that, we omitted them from the main paper. However, they are reported in the supplementary material for completeness.

The results for anomaly detection on MVTec LOCO are shown in Table~\ref{tb:results}. SALAD achieves the best score with a mean average AUROC of 96.1\%, beating the best previous method by a significant margin of $3.0$ percentage points. To enable future comparison, SALAD was also trained with composition maps from PSAD~\cite{psad}, which were obtained in a supervised manner. In this scenario, marked SALAD$^\dagger$, it outperforms all methods with supervised (or category-tuned) composition maps by a significant margin of $1.9$ percentage points. The results split by the anomaly type can be seen in Table~\ref{tb:results_split}. SALAD achieves both the highest logical anomaly detection and the highest structural anomaly detection result, suggesting that introducing the composition branch improves anomaly detection efficiency.

Table~\ref{tb:results_mvtec} shows the results on the structural anomaly datasets MVTec AD~\cite{mvtec} and Visa~\cite{visa}. SALAD achieves a state-of-the-art result with a mean average AUROC of 98.3\% over both datasets. SALAD outperforms the vast majority of logical anomaly detection methods, showing superior performance in scenarios with only structural anomalies despite not being specialised for this task.

Qualitative examples can be seen in Figure~\ref{fig:res}. SALAD produces accurate anomaly localisation even in hard near-distribution cases, with which previously proposed methods struggled (Columns 5, 8, 11 and 12). The extra successful detections are primarily due to the composition branch, and the presented anomalies mainly concern the image's composition. SALAD also detects all of the regions containing an anomaly as opposed to previous methods (Columns 1, 5 and 11). The better coverage comes from the composition branch, which detects different parts of the anomaly.

\section{Ablation study}

\begin{table*}[t]
\centering
\resizebox{0.7\textwidth}{!}{
    \begin{tabular}{llccc}
        \toprule
        \textbf{Group} & \textbf{Condition} & Det. Logical & Det. Struct. & Det. Avg\\
        \midrule
        \multirow{3}{*}{\textit{Full model}} & w/o Appearance branch & 95.6 \textcolor{blue}{(-0.9)} & 91.2 \textcolor{blue}{(-4.5)} & 93.4 \textcolor{blue}{(-2.7)} \\
        & w/o Composition branch & 93.0 \textcolor{blue}{(-3.5)} & 94.7 \textcolor{blue}{(-1.0)} & 93.8 \textcolor{blue}{(-2.3)}\\
        & w/o Stat branch & 93.6 \textcolor{blue}{(-2.9)} & 94.9 \textcolor{blue}{(-0.8)} & 94.2 \textcolor{blue}{(-1.9)}\\
        
        \midrule
        \multirow{3}{*}{\textit{Appearance branch}} & DSR~\cite{dsr} & 95.2 \textcolor{blue}{(-1.3)} & 93.7 \textcolor{blue}{(-2.0)}& 94.4 \textcolor{blue}{(-1.7)}\\
        & TransFusion~\cite{transfusion} & 95.3 \textcolor{blue}{(-1.2)} & 95.5 \textcolor{blue}{(-0.2)} & 95.4 \textcolor{blue}{(-0.7)}\\
        & DR{\AE}M~\cite{draem} & 94.8 \textcolor{blue}{(-1.7)} & 93.1 \textcolor{blue}{(-2.6)}& 94.0 \textcolor{blue}{(-2.1)}\\
        \midrule
        \multirow{3}{3cm}{\textit{Composition branch}} & w/o DR{\AE}M anomalies & 96.0 \textcolor{blue}{(-0.5)} & 95.2 \textcolor{blue}{(-0.5)} & 95.6 \textcolor{blue}{(-0.5)} \\
        & w/o component inpainting anomalies & 95.9 \textcolor{blue}{(-0.6)} & 95.6 \textcolor{blue}{(-0.1)} & 95.7 \textcolor{blue}{(-0.4)} \\
        & w/o component removal anomalies & 95.5 \textcolor{blue}{(-1.0)} & 95.1 \textcolor{blue}{(-0.6)} & 95.2 \textcolor{blue}{(-0.9)} \\
        \midrule
        \multirow{4}{3cm}{\textit{Global branch}} & Only Global Vector & 95.6 \textcolor{blue}{(-0.9)} & 95.6 \textcolor{blue}{(-0.1)} & 95.6 \textcolor{blue}{(-0.5)} \\
        & $g_\text{DINOv2}$ & 95.4 \textcolor{blue}{(-0.5)} & 93.5 \textcolor{blue}{(-2.4)} & 94.4 \textcolor{blue}{(-1.7)}\\ 
        & $g_\text{DINO}$ & 96.2 \textcolor{blue}{(-0.3)} & 94.3 \textcolor{blue}{(-1.4)} & 95.3 \textcolor{blue}{(-0.8)} \\
        & $g_\text{ResNet50}$ & 92.7 \textcolor{blue}{(-3.8)} & 92.8 \textcolor{blue}{(-3.0)} & 92.7 \textcolor{blue}{(-3.4)} \\ \midrule
        \multirow{3}{3cm}{\textit{Object composition \\ map generation}} & DINOv2 & 95.6 \textcolor{blue}{(-0.9)} & 95.4 \textcolor{blue}{(-0.3)} & 95.5 \textcolor{blue}{(-0.6)} \\
        & 4 clusters & 95.4 \textcolor{blue}{(-0.3)} & 95.4 \textcolor{blue}{(-0.8)} & 95.4 \textcolor{blue}{(-0.7)} \\ 
        & 8 clusters & 96.0 \textcolor{blue}{(-0.5)} & 95.8 \textcolor{blue}{(+0.1)} & 95.9 \textcolor{blue}{(-0.2)} \\ \midrule
        \textit{SALAD} & EfficientAD, DINO, 6 clusters & 96.5 & 95.7 & 96.1 \\
        \bottomrule
    \end{tabular}
}
\caption{Ablation study results. Results are reported for MVTec LOCO~\cite{mvtec-loco} in AUROC and separated by the type of anomalies. In the last column, the average for both types is reported. The difference from the base model is shown in \textcolor{blue}{blue}.}
\label{tb:ablation}
\end{table*}

Ablation experiments validating the contributions of SALAD are performed. Results are shown in Tables~\ref{tb:ablation} and \ref{tb:inf_speed}.

\noindent \textbf{Branch performance} To show each branch's overall importance and especially the composition branch's importance, we evaluated the model by excluding one branch at a time. Dropping the appearance branch leads to a $0.9$ p.\ p. drop in logical anomalies and a $4.5$ p.\ p.\ drop in structural anomalies. Dropping the composition branch results in a $3.5$ p.\ p.\ drop in performance with logical anomalies and a $1.0$ p.\ p.\ drop with structural anomalies. Dropping the stat branch results in the lowest overall performance drop with an overall drop of $1.8$ p.\ p. While not using the composition branch would still achieve SOTA results, it wouldn't improve them significantly. This confirms that modelling the composition map distribution will improve logical anomaly detection.

\noindent \textbf{Choice of the architecture} To show the generality of the proposed framework, we exchanged the appearance branch with three other state-of-the-art models: DSR~\cite{dsr}, TransFusion~\cite{transfusion} and DR{\AE}M~\cite{simplenet}. To maintain a unified evaluation process, we disabled the centre-cropping for TransFusion. No model selection strategy or parameter tuning was performed for each model. Using all three architectures for the appearance branch produced state-of-the-art results. TransFusion performs better than the other two due to better overall performance with structural anomalies, where the composition and the global branch struggle. The results still show robustness to the choice of the appearance branch.

\noindent \textbf{Different synthetic anomaly generation strategies} Three synthetic anomaly generation techniques are used during training - DR{\AE}M anomaly, component inpainting and component removal. Each strategy was removed from training to verify its contribution to the overall performance. Removing each strategy resulted in a drop in performance. The highest drop is seen by removing the component removal strategy (0.9 p.\ p.) and the lowest when we remove the component inpainting strategy (0.4 p.\ p.). The results show the contribution and necessity of each strategy. 

\noindent \textbf{Different global representation} To investigate the improvement of the global representation, we exchanged our global representation with the one from PUAD~\cite{puad}, that is, using only the global mean vector. The performance drops by 0.9 p.\ p. in logical anomalies and by 0.1 p.\ p.\ in structural anomalies. The drop is due to the lack of spatial information inside the global representation. The results indicate that our representation does indeed improve the results.

\noindent \textbf{Different feature representation for the global representation} To verify the effectiveness of EfficientAD's feature extractor for the global representation, we exchanged it with a few different high-performing feature extractors -- DINOv2~\cite{dinov2}, DINO~\cite{dino} and ResNet50~\cite{resnet}. The performance drops the least (0.8 p.\ p.) when using DINO and the most using ResNet50 (3.4 p.\ p.). We hypothesise this is due to the subpar representation of ResNet50 features. However, it shows the importance of choosing the right feature extractor for the global representation.

\noindent \textbf{Different feature extractor for the composition map generation} To investigate the importance of the feature extractor in the object composition map generation, we exchanged DINO with DINOv2~\cite{dinov2}. The performance drops by 0.9 p.\ p.\ in logical anomalies and by 0.3 p.\ p.\ in structural anomalies. The drop is due to consistent small mistakes in the generated composition maps made by DINOv2. Some examples are in the Supplementary material. Nevertheless, the results suggest that the choice of feature extractor for the composition map generation is robust.

\noindent \textbf{Different number of clusters} To show the robustness of the cluster number parameter, we also evaluated our model for $K=4$ and $K=8$. Having 4 clusters results in a 0.6 p.\ p. drop in overall performance, and having 8 clusters results in a 0.2 p.\ p. drop in overall performance. These results suggest that the results are robust when $K$ gets high enough. If $K$ is too low, the results are lower. Qualitative examples and the results for other values are in the Supplementary material.

\begin{table}
\centering
\resizebox{\columnwidth}{!}{
\begin{tabular}{lcccc} \toprule
\textbf{Method} & DR{\AE}M~\cite{draem} & Patchcore~\cite{patchcore}  & EfficientAD~\cite{efficientad} & \textit{SALAD} \\ \midrule
Inference [ms] & \bm2{52.6} & 224.4 & \bm1{6.2} & \bm3{64.6} \\ \bottomrule
\end{tabular}
}
\caption{Results for average inference time of a single sample with NVIDIA A100 GPU. Inference times are reported in milliseconds.}
\vspace{-0.6cm}
\label{tb:inf_speed}
\end{table}

\noindent \textbf{Inference Speed and Computational Complexity} The inference speed can be seen in Table~\ref{tb:inf_speed}. SALAD is faster than Patchcore~\cite{patchcore} and lags slightly behind DR{\AE}M~\cite{draem} and EfficientAD~\cite{efficientad}. SALAD could be further optimised for speed by successfully parallelising each branch. SALAD requires approximately 1.5 hours to train on a single A100 GPU and has $65.1$ million parameters.

\section{Conclusion}
A novel model for logical anomaly detection, SALAD, is proposed. Unlike recent methods, SALAD explicitly models object composition information by introducing a novel discriminatively trained composition branch. For this purpose, it introduces a novel automatic composition map generation strategy and an anomaly simulation process, facilitating discriminative training. SALAD achieves a new state-of-the-art of 96.1\% AUROC on the MVTec LOCO Dataset, outperforming all previous methods by a significant margin of 3.0 percentage points. Furthermore, SALAD also performs very well on datasets with only structural anomalies, achieving 98.9\% on MVTec AD and 97.9\% on VisA. Further interaction between branches in the architecture may improve performance and is a good avenue for future research. The results indicate that explicit composition distribution modelling is also a viable future research direction.

\noindent \textbf{Acknowledgements} This work was in part supported by the ARIS research project MUXAD (J2-60055), research programme P2-0214 and the supercomputing network SLING (ARNES, EuroHPC Vega).

{
    \small
    \bibliographystyle{ieeenat_fullname}
    \bibliography{main}
}

\clearpage
\clearpage
\setcounter{page}{1}
\maketitlesupplementary

\newpage

\appendix
\setcounter{table}{0}
\setcounter{figure}{0}

This supplementary material includes additional information and visualisations. More specifically, we ablate the object composition map generation and add a further experiment to verify the importance of each branch. Ultimately, we add localisation results for MVTec LOCO and more qualitative examples.

\section{Limitations and Failure Cases}

\begin{figure*}
    \centering
    \includegraphics[width=\textwidth]{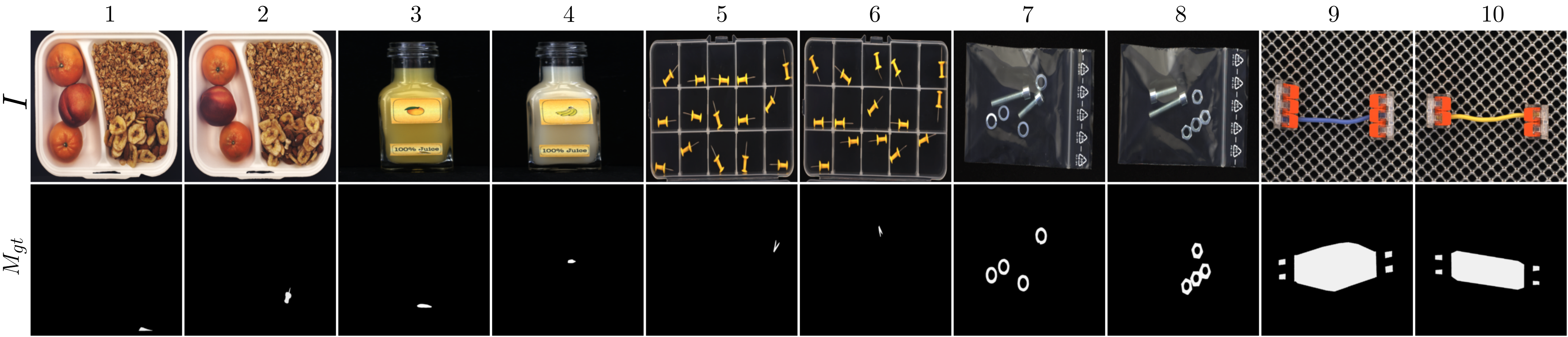}
    \caption{Failure case results. In all of the cases, SALAD produces a very low anomaly score. Most of the cases also represent near-distribution logical and structural anomalies.}
    \label{fig:fail}
\end{figure*}

Composition map creation depends on the performance of SAM-HQ and DINO, although they perform very well across diverse datasets. In the future, this can even be improved with stronger models (e.g. Perception Encoder). Additionally, there are also some cases (some are depicted in Figure~\ref{fig:fail}) in which SALAD fails to detect anomalies. SALAD mostly fails on extremely near-distribution structural (Columns 1-6) and logical anomalies (Columns 7-10). Architectural improvements to the compositional and appearance branch might improve this.

\section{Differences from other methods utilising composition maps}

Currently, there are three different methods using composition maps -- ComAD~\cite{comad}, CSAD~\cite{csad} and PSAD~\cite{psad}. SALAD's biggest difference from all three is the introduction of a specialised composition branch. This means SALAD is directly trained on the composition maps in contrast to the other three. Additionally, CSAD and PSAD require extra category-specific information, either via hand-labelled samples or via category-specific composition map procedures. SALAD performs all of this automatically without any additional information. ComAD produces composition maps of low quality, whilst SALAD produces high-quality maps.

\section{Object composition map generation ablation}

\begin{figure*}
    \centering
    \includegraphics[width=0.8\textwidth]{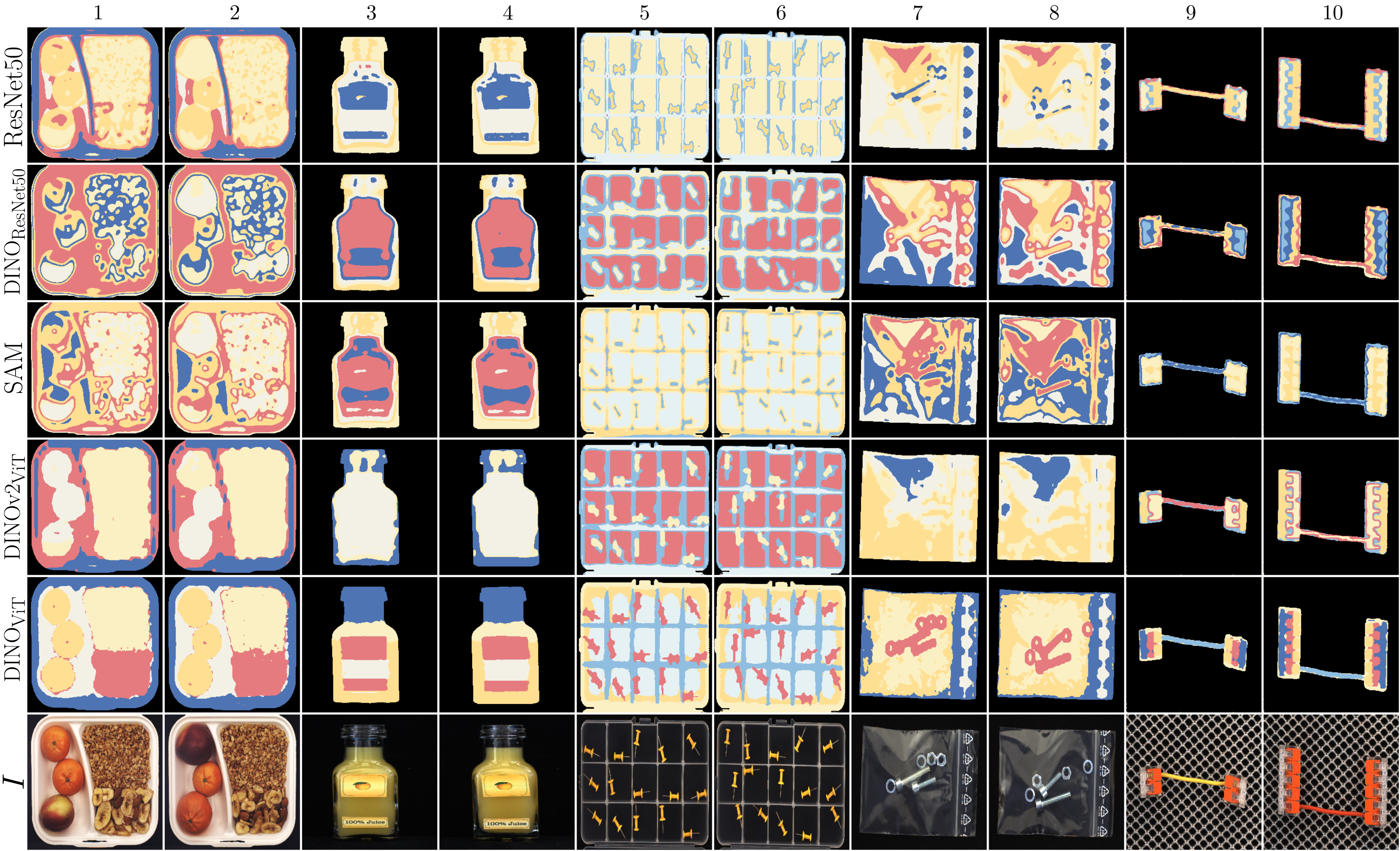}
    \caption{Qualitative comparison of the feature clusters $C_{feat}$ produced by 5 different feature extractors: ResNet50~\cite{resnet}, ResNet50 DINO~\cite{dino}, SAM~\cite{sam_hq} and ViT DINO~\cite{dino}. In the bottom row, the original image $I$ is shown. It can be observed that both ViT DINOv2 and ViT DINO separate the objects effectively (e.g. Columns 5 and 6), while other feature extractors face problems (e.g. Columns 3 and 4).}
    \label{fig:sup_feat_clust}
\end{figure*}

\begin{figure*}
    \centering
    \includegraphics[width=0.8\textwidth]{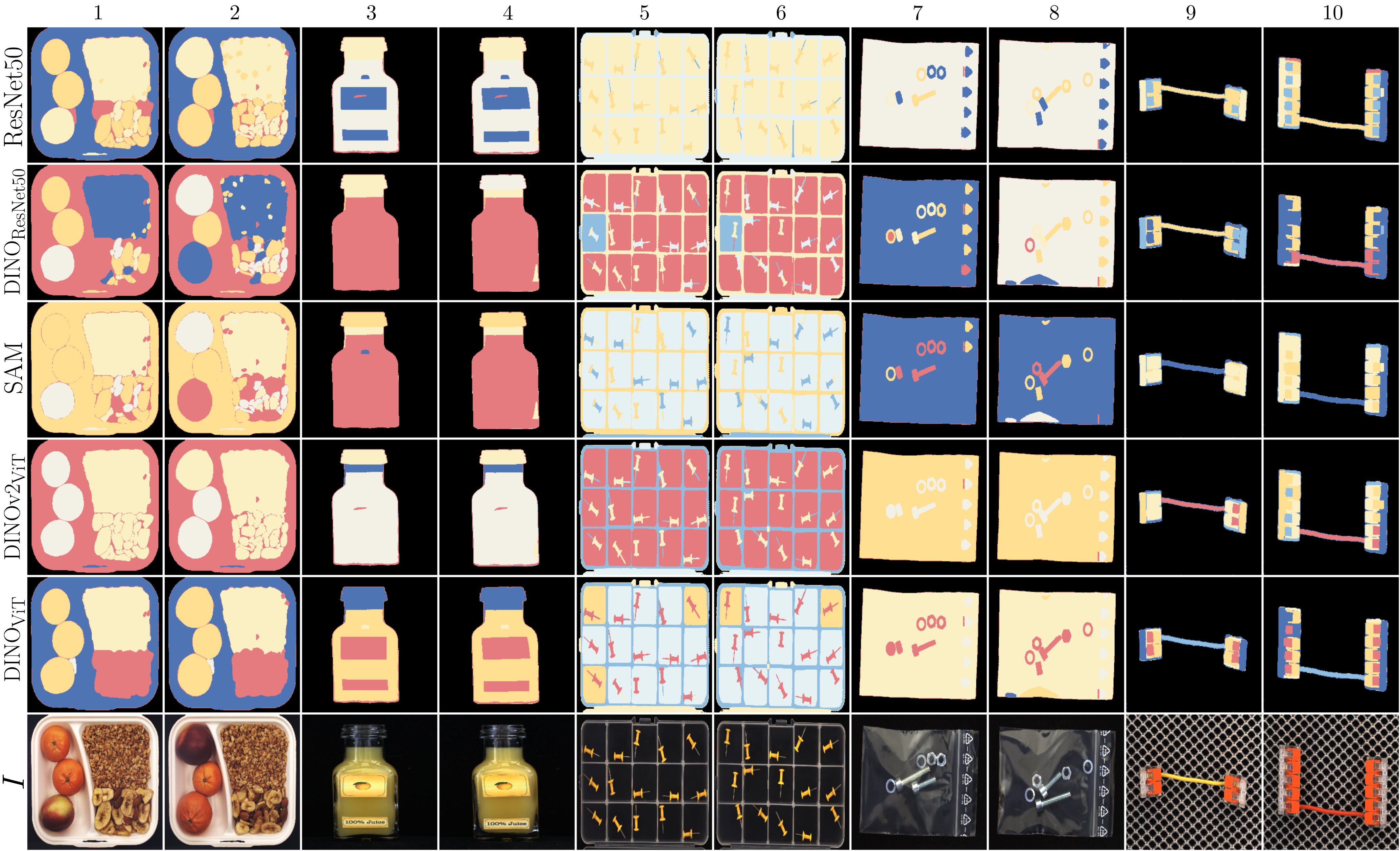}
    \caption{Qualitative comparison of the pseudo-labels $C_{pseudo}$ produced by 5 different feature extractors: ResNet50~\cite{resnet}, ResNet50 DINO~\cite{dino}, SAM~\cite{sam_hq} and ViT DINO~\cite{dino}. In the bottom row, the original image $I$ is shown. Most methods do not face class mismatches and loss of detail except for ResNet50, DINO, and SAM (e.g. Columns 7 and 8).}
    \label{fig:sup_noisy_cmaps}
\end{figure*}

\begin{figure*}
    \centering
    \includegraphics[width=0.8\textwidth]{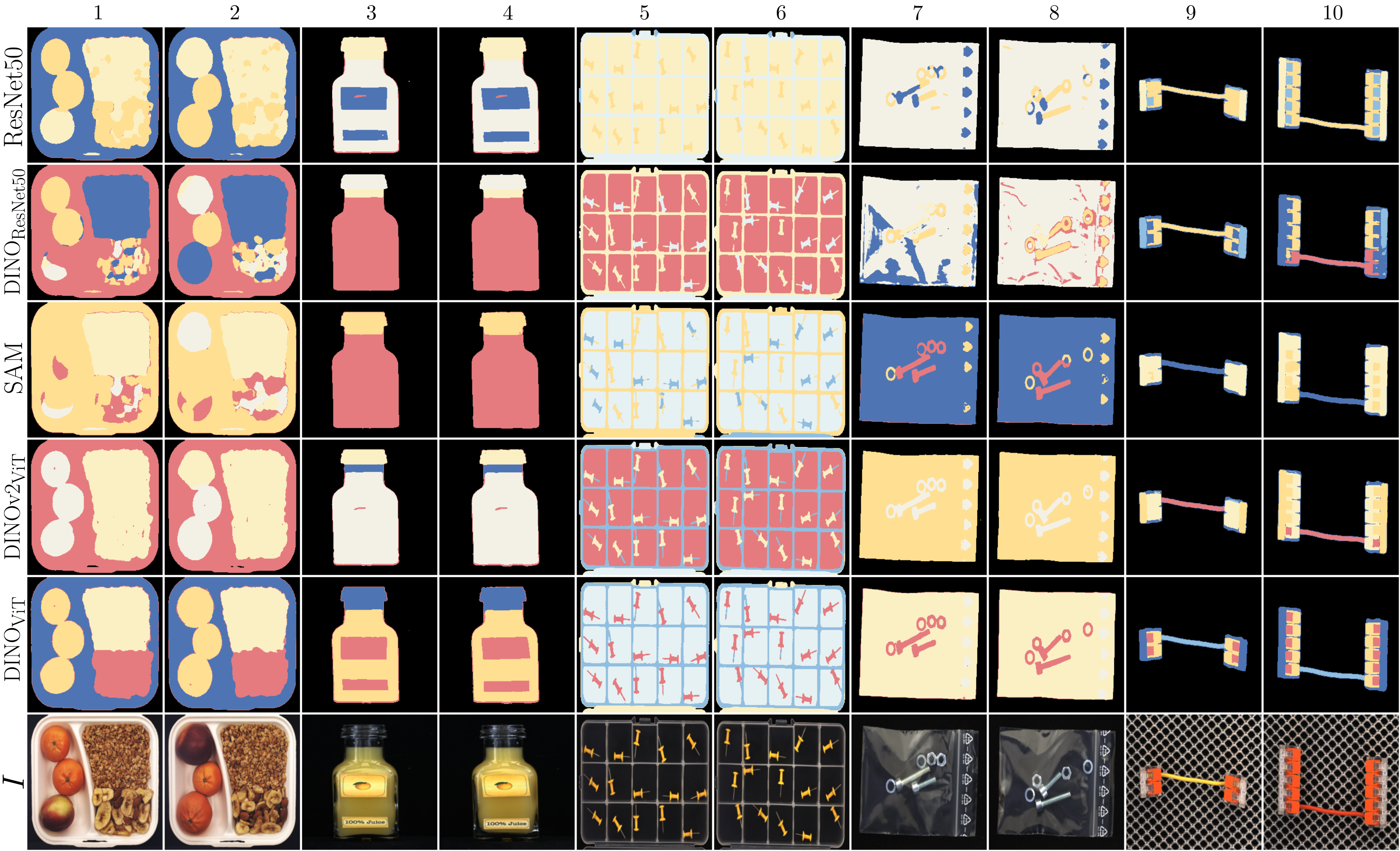}
    \caption{Qualitative comparison of the composition maps $C$ produced by 5 different feature extractors: ResNet50~\cite{resnet}, ResNet50 DINO~\cite{dino}, SAM~\cite{sam_hq} and ViT DINO~\cite{dino}. In the bottom row, the original image $I$ is shown. While ViT DINO and ViT DINOv2 can generalise effectively, other methods face problems (e.g. Columns 7 and 8).}
    \label{fig:sup_camps}
\end{figure*}

This section compares the design choices for the object composition map generation. First, we examine the effect of using a different feature extractor than DINO~\cite{dino}. Then, we examine the importance of the number of clusters parameter.

\noindent \textbf{Different feature extractor} To evaluate the choice of feature extractor in component map generation, we replaced the original with other standard feature extractors: ResNet50~\cite{resnet}, ResNet50 DINO~\cite{dino}, SAM~\cite{sam_hq}, ViT DINOv2~\cite{dinov2}. Their performance is qualitatively evaluated by comparing feature clusters $C_{feat}$, pseudo labels $C_{pseudo}$, and final composition maps $C$. Figure~\ref{fig:sup_feat_clust} depicts that ResNet50, ViT DINOv2, and ViT DINO cluster features effectively, discriminating similar objects (e.g., Columns 5 and 6). In contrast, ResNet, DINO and SAM yield poor clusters, as seen in Columns 3 and 4. This pattern continues with pseudo labels in Figure~\ref{fig:sup_noisy_cmaps}, where ResNet DINO and SAM exhibit loss of detail and class mismatches (Columns 7 and 8). Due to noisy pseudo labels, the lightweight semantic segmentation model struggles with generalisation (Figure~\ref{fig:sup_camps}, Columns 7 and 8). Consequently, we evaluated downstream anomaly detection performance only for ViT DINO~\cite{dino} and ViT DINOv2~\cite{dinov2}, with results detailed in the main paper.

\begin{figure*}
    \centering
    \includegraphics[width=0.8\textwidth]{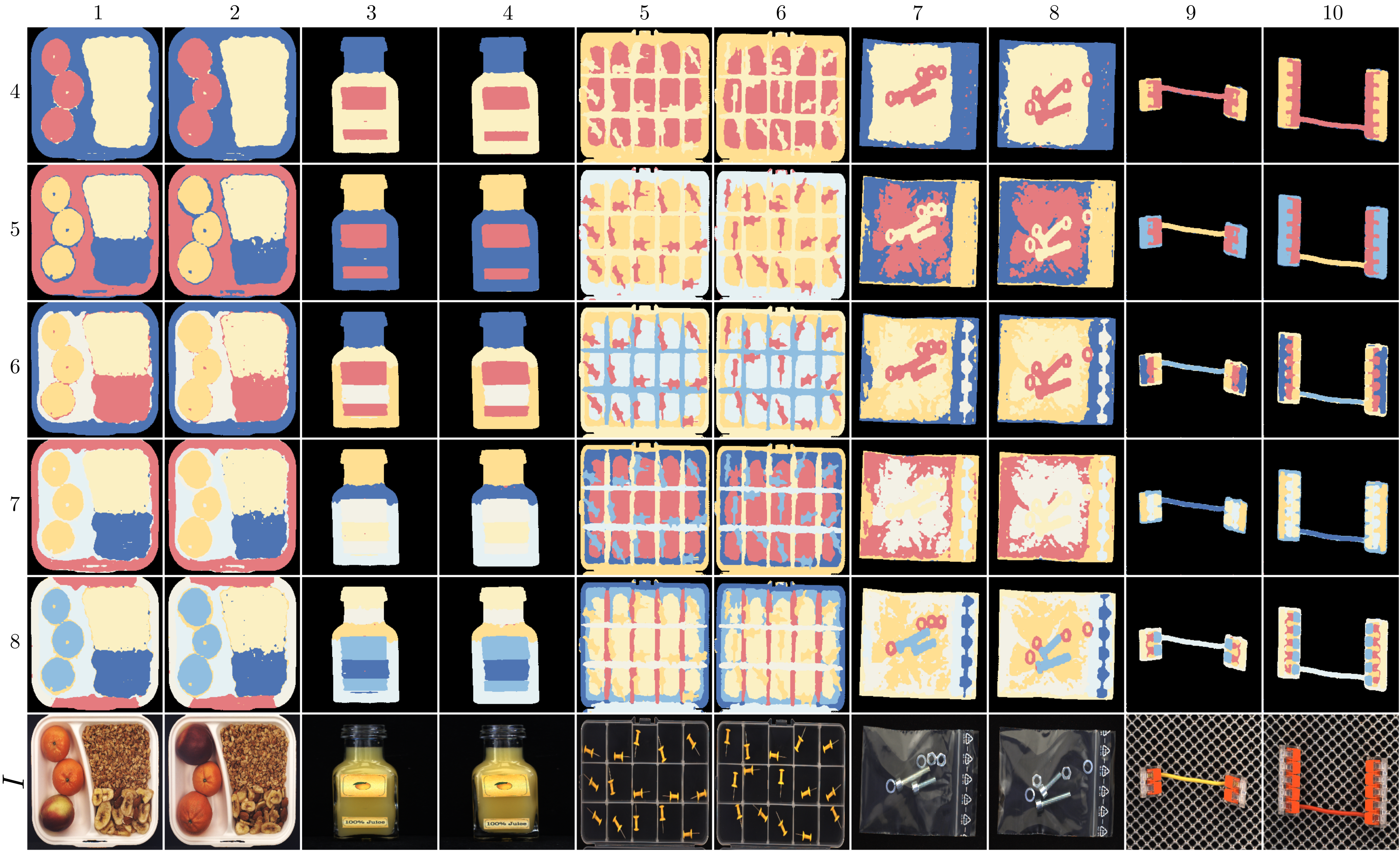}
    \caption{Qualitative comparison of the feature clusters $C_{feat}$ produced by different numbers of clusters $K$ (from 4 to 8). In the bottom row, the original image $I$ is shown.}
    \label{fig:sup_feat_clust_num}
\end{figure*}

\begin{figure*}
    \centering
    \includegraphics[width=0.8\textwidth]{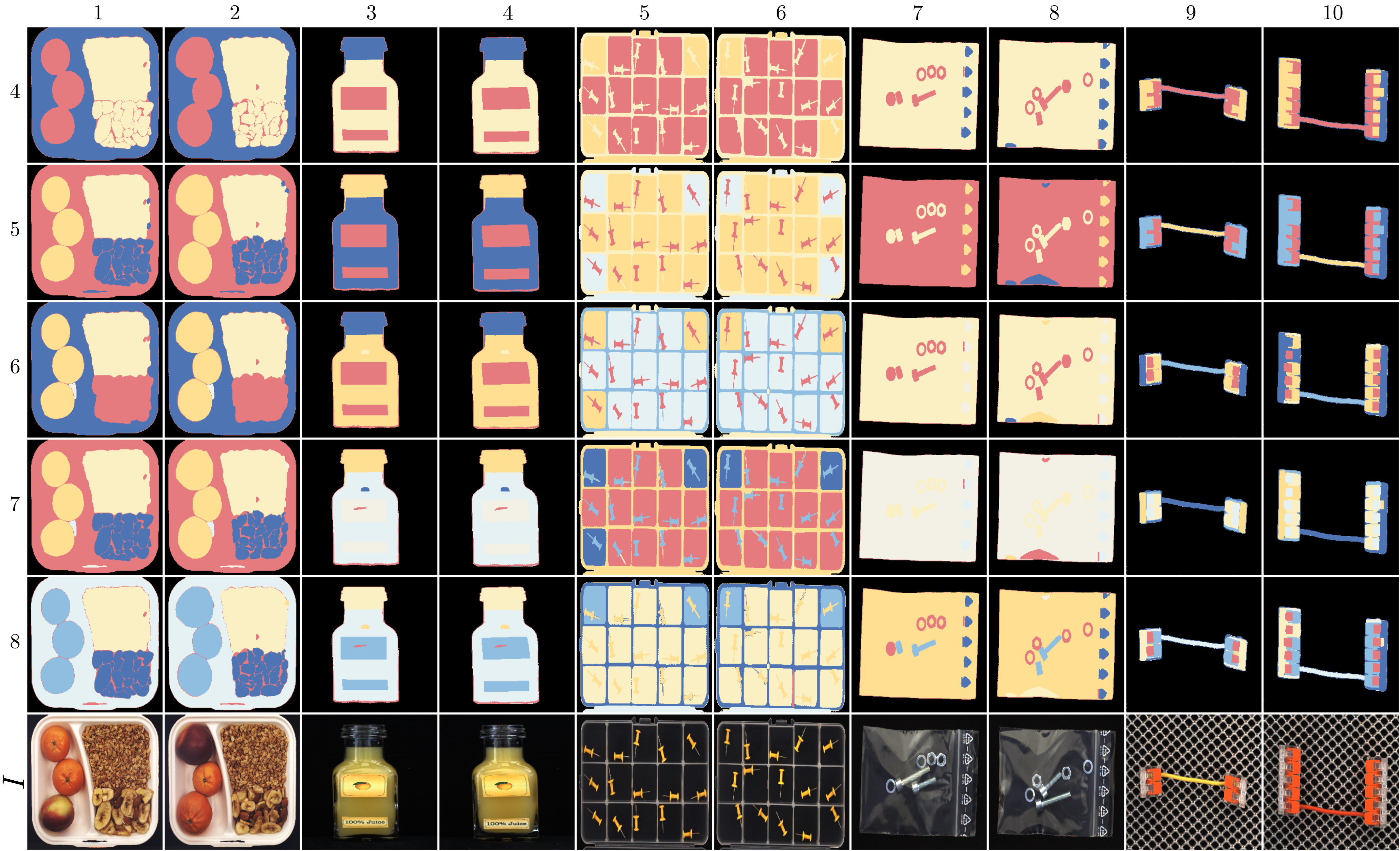}
    \caption{Qualitative comparison of the pseudo-labels $C_{pseudo}$ produced by different numbers of clusters $K$ (from 4 to 8). }
    \label{fig:sup_noisy_cmaps_clust_num}
\end{figure*}

\begin{figure*}
    \centering
    \includegraphics[width=0.8\textwidth]{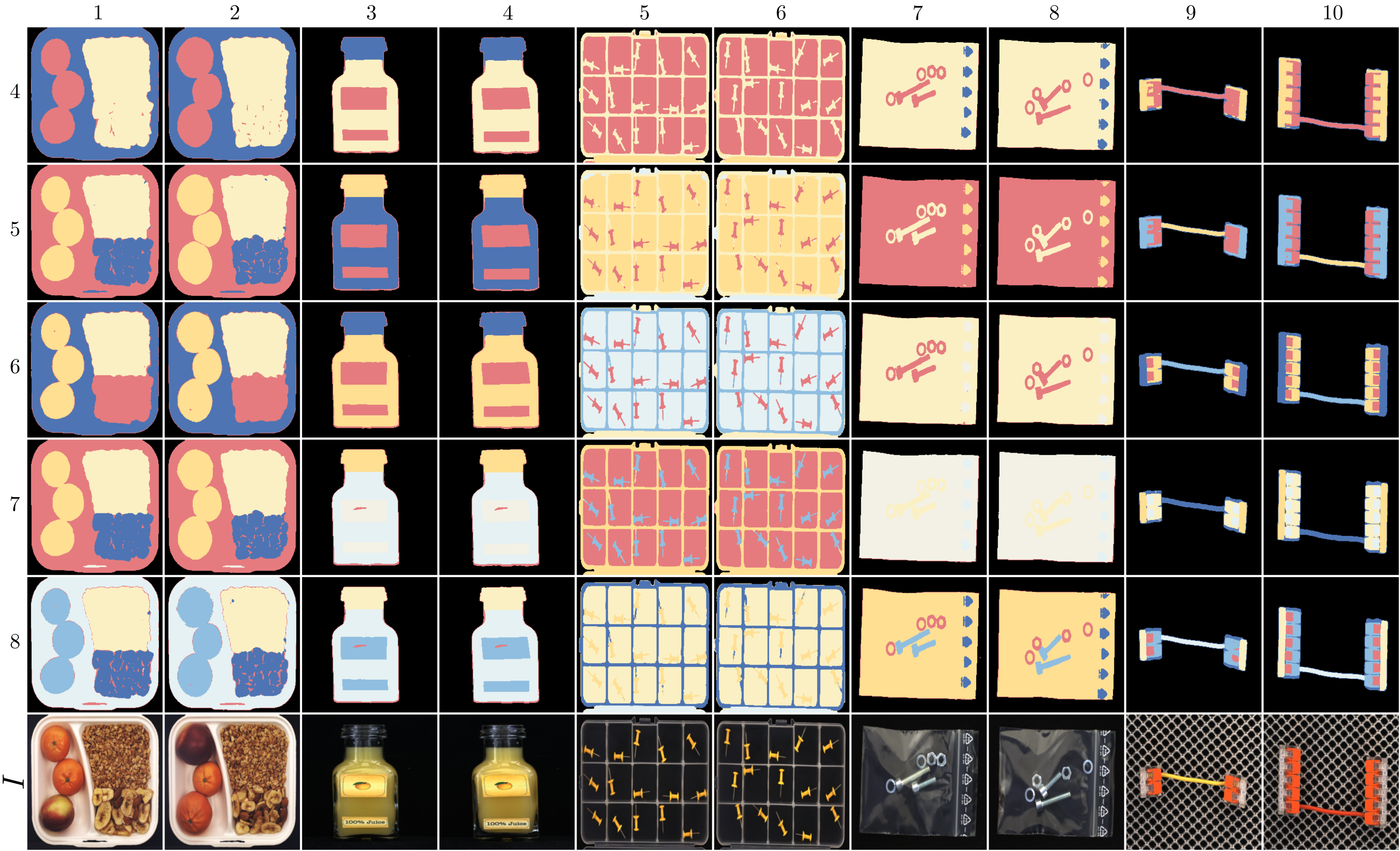}
    \caption{Qualitative comparison of the composition maps $C$ produced by different numbers of clusters $K$ (from 4 to 8).}
    \label{fig:sup_camps_clust_num}
\end{figure*}

\begin{figure*}
    \centering
    \includegraphics[width=0.5\textwidth]{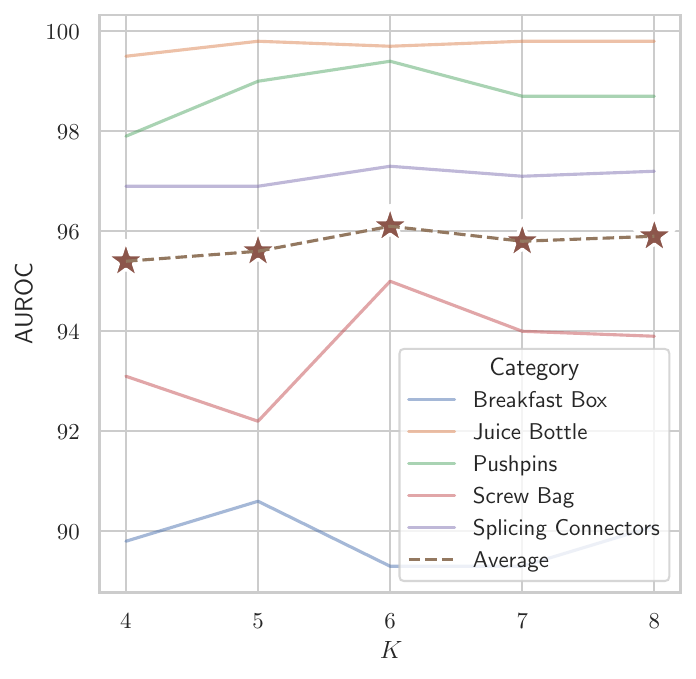}
    \caption{Anomaly detection performance on MVTec LOCO under different values for $K$ in the object composition map generation. The default settings for $K$ is $6$.}
    \label{fig:ablation_stat_clust}
\end{figure*}

\noindent \textbf{Different number of clusters} To investigate the importance of the number of clusters during composition map generation, we qualitatively and quantitatively assessed the output composition maps. More specifically, we checked for different values of $K$ ranging from 4 to 8. We qualitatively assessed the feature clusters, pseudo-labels and the generated composition maps. The results for different stages in the pipeline can be seen in Figure~\ref{fig:sup_feat_clust_num}, Figure~\ref{fig:sup_noisy_cmaps_clust_num} and Figure~\ref{fig:sup_camps_clust_num}. From the Figures, it can be seen that there are no significant differences, especially with the final composition maps. This would suggest that the choice of the number of clusters is robust (once it is high enough). In Figure~\ref{fig:ablation_stat_clust}, the effect of this parameter on downstream anomaly detection is depicted. All values are above the current state-of-the-art, suggesting that the parameter choice is robust.

\section{Branch importance}

\begin{table}[t]
\centering
\resizebox{\columnwidth}{!}{
\begin{tabular}{lccc}
\toprule
\textbf{Condition} & Det. Logical & Det. Struct. & Det. Avg\\
\midrule
Only Appearance branch & 87.5 \textcolor{blue}{(-9.0)} & 94.1 \textcolor{blue}{(-1.6)} & 90.8 \textcolor{blue}{(-5.3)}\\
Only Composition branch & 88.1 \textcolor{blue}{(-8.4)} & 82.8 \textcolor{blue}{(-12.9)} & 85.4 \textcolor{blue}{(-10.7)}\\
Only Global branch & 90.8 \textcolor{blue}{(-5.7)} & 87.3 \textcolor{blue}{(-8.4)} & 89.1 \textcolor{blue}{(-8.1)}\\ \midrule
\textit{SALAD} & 96.5 & 95.7 & 96.1 \\
\bottomrule
\end{tabular}
}

\caption{Branch importance is evaluated with the downstream performance in Anomaly detection on the MVTec LOCO dataset~\cite{mvtec-loco} (results are presented in AUROC). The importance is evaluated by using only one branch. The results are categorised by anomaly type, and the overall average detection rate is reported in the final column. The performance difference relative to the base model is highlighted in \textcolor{blue}{blue}.}
\label{tb:ablation_sup}
\end{table}

\begin{table*}[t]
\centering
    \begin{tabular}{lcccccc}
\toprule
\textbf{Branch} & Breakfast box & Juice bottle & Pushpins & Screw bag & Splicing conn. & \textit{Average} \\ \midrule 
Only Appearance Branch & 85.7 & 96.9 & 96.8 & 77.9 & 96.6 & 90.8  \\
Only Composition Branch & 77.1 & 87.0 & 87.7 & 88.2 & 86.2 & 85.4  \\
Only Global Branch & 82.2 & 97.7 & 91.8 & 86.3 & 87.3 & 89.1  \\
 \bottomrule 
    \end{tabular}
\caption{Anomaly detection (AUROC) for each branch on MVTec LOCO~\cite{mvtec-loco}.}
\label{tb:results_branch} 
\end{table*}

To further show the overall importance of each branch, we evaluated the model by using one branch at a time. The results can be seen in Table~\ref{tb:ablation_sup} and in Table~\ref{tb:results_branch}. Using only the appearance branch leads to a drop in performance of $9.0$ percentage points (p.\ p.) for logical anomalies and $1.6$ p.\ p.\ for structural anomalies. Using only the composition branch leads to a drop of $8.4$ p.\ p.\ on logical anomalies and a $12.9$ p.\ p.\ drop for structural anomalies. By solely using the global branch, the performance drops $5.7$ p.\ p.\ for logical anomalies and $8.4$ p.\ p.\ for structural anomalies. The results show that the branches complement each other, especially with logical anomalies. 

\section{Localisation results for MVTec LOCO}

\begin{figure*}[!h]
    \centering
    \includegraphics[width=1\textwidth]{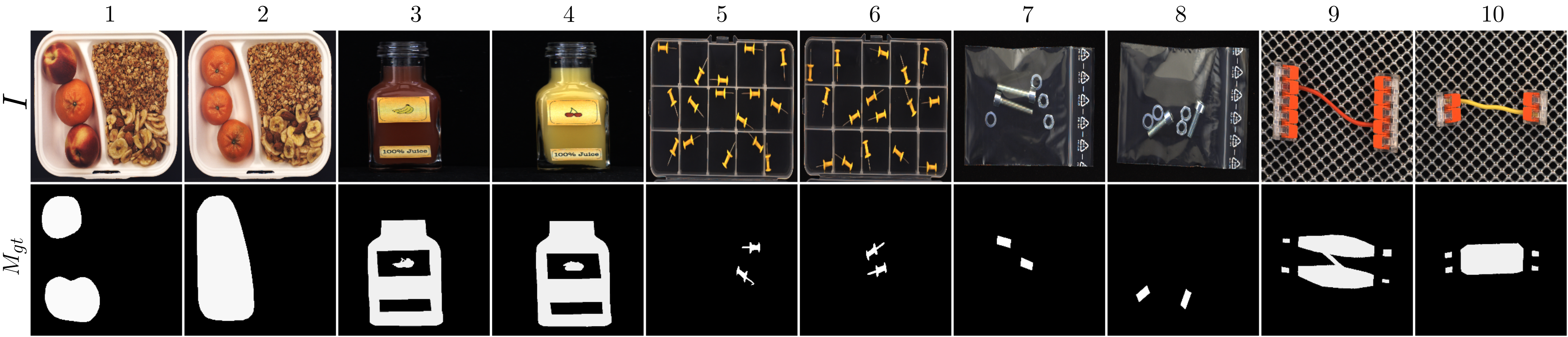}
    \caption{Examples of problematic pixel-level ground truths ($M_{gt}$) and their corresponding images ($I$) in MVTec LOCO~\cite{mvtec-loco} show issues with how annotations are done. The ground truths are designed to include all possible solutions, which causes ambiguity. For example, in Column 7, there are two long screws instead of one long screw and one short screw as expected. The annotation requires marking both long screws, even though marking one would still be a correct interpretation of the anomaly. This approach unfairly lowers the scores of methods that label only one screw, even if their prediction makes sense.}
    \label{fig:loco_problems}
\end{figure*}

Following recent literature~\cite{efficientad,slsg}, the AUsPRO Metric~\cite{mvtec-loco} is used to evaluate the localisation performance. Again, it is important to highlight that most concurrent works~\cite{psad, csad, puad, comad, sinbad} strayed away from reporting these results due to the ambiguity in pixel-level ground truths in images containing a logical anomaly. Some such cases are depicted in Figure~\ref{fig:loco_problems}. The localisation results on MVTec LOCO are given in Table~\ref{tb:results_loc}. SALAD achieves the second-highest result with an AUsPRO of $68.7$\%.

\begin{table*}[h]
\centering
\resizebox{\textwidth}{!}{
    \begin{tabular}{lccccccccc}
\toprule
\textbf{Category} & SimpleNet~\cite{simplenet} & DR{\AE}M~\cite{draem} & TransFusion~\cite{transfusion} & DSR~\cite{dsr} & Patchcore~\cite{patchcore} & SLSG~\cite{slsg} & EfficientAD~\cite{efficientad} & \textit{SALAD} \\ \midrule 
Breakfast box & 38.8 & 49.9 & \bm3{53.5} & 49.9 & 46.6 & \bm1{65.9} & \bm2{60.4} & 49.1 \\
Juice bottle & 43.9 & 80.0 & \bm2{90.1} & \bm3{86.8} & 41.2 & 82.0 & \bm1{93.4} & 81.5 \\
Pushpins & 27.2 & 49.3 & 51.9 & 59.1 & 31.4 & \bm1{74.4} & \bm3{62.3} & \bm2{73.5} \\
Screw bag & \bm1{66.0} & 49.0 & 39.3 & 37.9 & 48.1 & 47.2 & \bm2{64.4} & \bm3{58.4} \\
Splicing connectors & 36.9 & \bm3{67.3} & 67.0 & 58.6 & 31.3 & 66.9 & \bm2{73.3} & \bm1{81.2} \\ \midrule
\textit{Average} & 36.3 & 59.1 & 60.4 & 58.5 & 39.7 & \bm3{67.3} & \bm1{69.4} & \bm2{68.7} \\ 
 \bottomrule 
    \end{tabular}
}
\caption{Anomaly localization (AUsPRO) on MVTec LOCO~\cite{mvtec-loco}.}
\label{tb:results_loc}
\end{table*}

\section{Additional qualitative results}

\begin{figure*}
    \centering
    \includegraphics[width=1\textwidth]{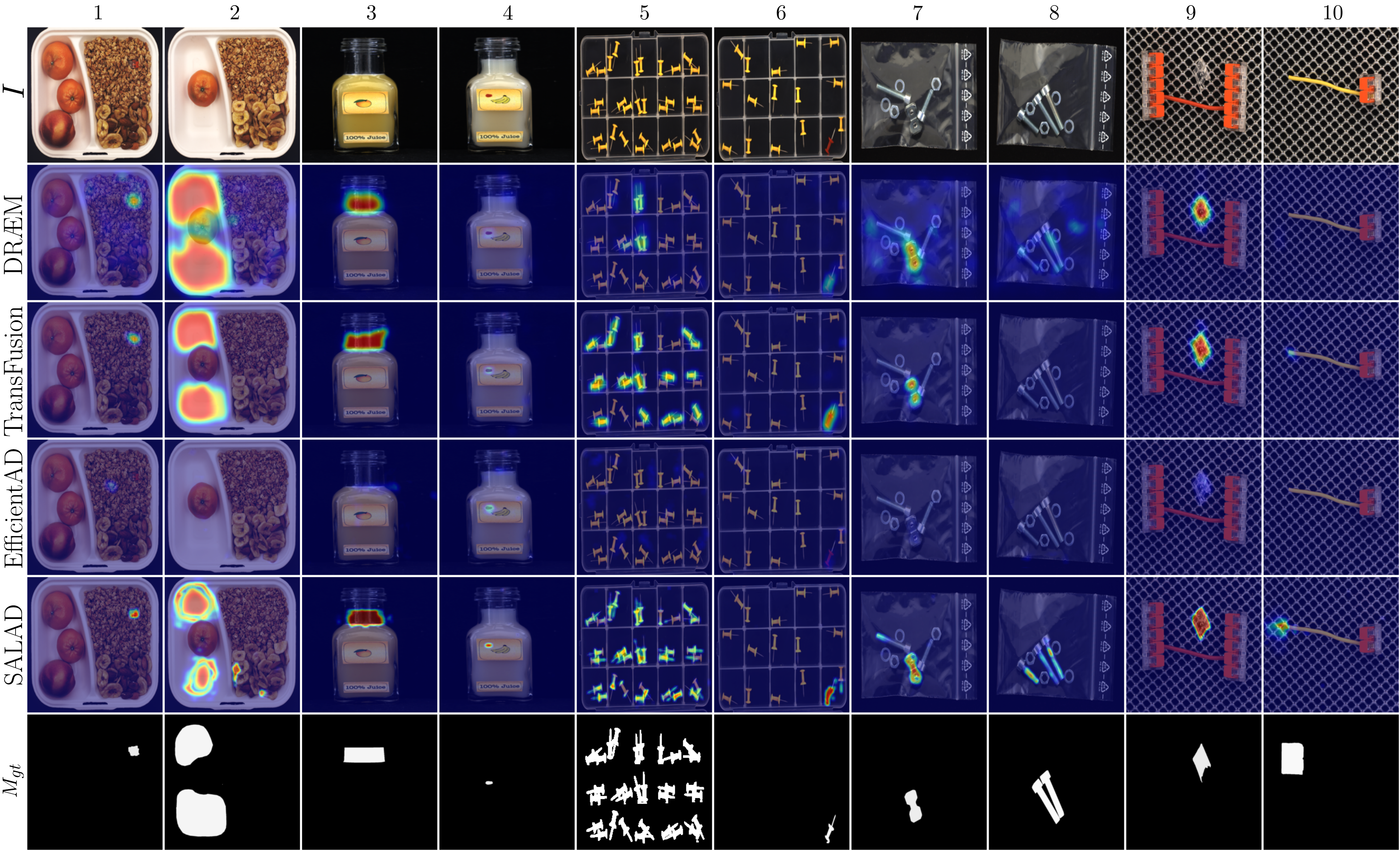}
    \caption{Qualitative comparison of the anomaly segmentation masks produced by SALAD and three other state-of-the-art methods on MVTec LOCO. In the first row, the image is shown. In the next four rows, the anomaly segmentations produced by DR{\AE}M~\cite{draem}, TransFusion~\cite{transfusion}, EfficientAD~\cite{efficientad} and SALAD are depicted, and in the last row, the ground truth mask is shown. For SALAD, we visualised the sum of $A_a$ and $A_c$.}
    \label{fig:sup_mask_loco}
\end{figure*}

\begin{figure*}
    \centering
    \includegraphics[width=0.6\textwidth]{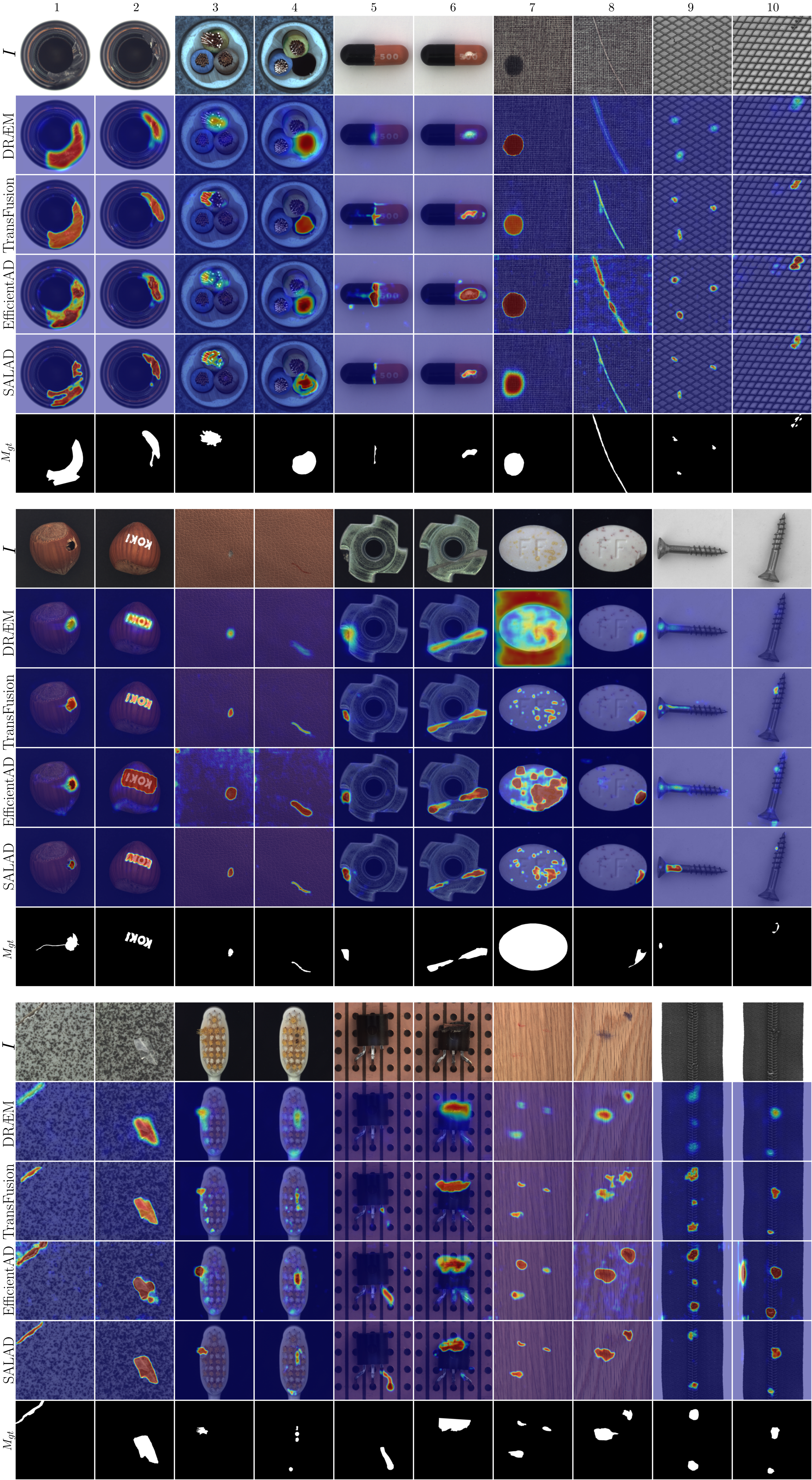}
    \caption{Qualitative comparison of the anomaly segmentation masks produced by SALAD and three other state-of-the-art methods on MVTec AD. In the first row, the image is shown. In the next four rows, the anomaly maps produced by DR{\AE}M~\cite{draem}, TransFusion~\cite{transfusion}, EfficientAD~\cite{efficientad} and SALAD are depicted, and in the last row, the ground truth mask is shown.}
    \label{fig:sup_mask_mvtec}
\end{figure*}

In this section, we provide additional qualitative mask comparisons to the state-of-the-art models DR{\AE}M~\cite{draem}, TransFusion~\cite{transfusion} and EfficientAD~\cite{efficientad}. The comparisons can be seen in Figure~\ref{fig:sup_mask_loco} and Figure~\ref{fig:sup_mask_mvtec}. SALAD can detect more near-distribution and harder anomalies compared to previous state-of-the-art methods.

\end{document}